\documentclass[acmlarge,screen,colorlinks,nonacm]{acmart}

\AtBeginDocument{}

\usepackage{hyperref}
\hypersetup{
colorlinks,
linkcolor=blue,
urlcolor=blue,
}

\usepackage{xspace}
\usepackage{balance}
\usepackage{amsmath}
\usepackage{multirow}
\usepackage{colortbl}
\usepackage{caption}
\usepackage[labelformat = parens, labelsep = space, textfont=small]{subcaption}

\usepackage{enumitem}
\usepackage[most]{tcolorbox} \usepackage{arydshln}
\usepackage{tabularx}

\newcommand{\sysname}{\textsc{CleAR}\xspace}

\definecolor{pro_green}{rgb}{0.0, 0.66, 0.47}
\definecolor{overleaf_green}{rgb}{0.08, 0.54, 0.02}

\newcommand{\1}{{\em (i)}}
\newcommand{\2}{{\em (ii)}}
\newcommand{\3}{{\em (iii)}}

\newcommand{\para }[1]{\noindent  {\bf #1}}
\setlist[itemize]{leftmargin=.12in}

\begin{document}

\title{\textsc{CleAR}: Robust Context-Guided Generative Lighting Estimation for Mobile Augmented Reality}

\author{Yiqin Zhao}
\orcid{0000-0003-1044-4732}
\affiliation{\institution{Worcester Polytechnic Institute}
 \streetaddress{100 Institute Road}
 \city{Worcester}
 \state{MA}
 \country{USA}
}
\email{yzhao11@wpi.edu}

\author{Mallesham Dasari}
\orcid{0000-0002-8855-036X}
\affiliation{\institution{Northeastern University}
  \streetaddress{360 Huntington Ave}
  \city{Boston}
  \state{MA}
  \country{USA}
}
\email{m.dasari@northeastern.edu}

\author{Tian Guo}
\orcid{0000-0003-0060-2266}
\affiliation{\institution{Worcester Polytechnic Institute}
 \streetaddress{100 Institute Road}
 \city{Worcester}
 \state{MA}
 \country{USA}
 }
\email{tian@wpi.edu}

\renewcommand{\shortauthors}{Zhao et al.}

\begin{abstract}

High-quality environment lighting is essential for creating immersive mobile augmented reality (AR) experiences.
However, achieving visually coherent estimation for mobile AR is challenging due to several key limitations in AR device sensing capabilities, including low camera FoV and limited pixel dynamic ranges.
Recent advancements in generative AI, which can generate high-quality images from different types of prompts, including texts and images, present a potential solution for high-quality lighting estimation.
Still, to effectively use generative image diffusion models, we must address two key limitations of \emph{content quality} and \emph{slow inference}.
In this work, we design and implement a generative lighting estimation system called \sysname that can produce high-quality, diverse environment maps in the format of $360^{\circ}$ HDR images.
Specifically, we design a two-step generation pipeline guided by AR environment context data to ensure the output aligns with the physical environment's visual context and color appearance.
To improve the estimation robustness under different lighting conditions, we design a real-time refinement component to adjust lighting estimation results on AR devices.
To train and test our generative models, we curate a large-scale environment lighting estimation dataset with diverse lighting conditions.
Through a combination of quantitative and qualitative evaluations, we show that \sysname outperforms state-of-the-art lighting estimation methods on both estimation accuracy, latency, and robustness, and is rated by 31 participants as producing better renderings for most virtual objects.
For example, \sysname achieves 51\% to 56\% accuracy improvement on virtual object renderings across objects of three distinctive types of materials and reflective properties.
\sysname produces lighting estimates of comparable or better quality in just 3.2 seconds---over 110X faster than state-of-the-art methods.
Moreover, \sysname supports real-time refinement of lighting estimation results, ensuring robust and timely updates for AR applications.

\end{abstract}



\maketitle

\section{Introduction}
\label{sec:introduction}

As new augmented reality (AR) hardware and software enter consumer markets, mobile AR technologies have positively impacted various industries, including e-commerce, education, and engineering~\cite{chylinski2020augmented,rauschnabel2019augmented}.
The growing public adoption of AR technologies demands new standards for content quality and application user experiences, particularly emphasizing the need for \textit{visual coherency} between virtual and physical content to ensure high-quality user experiences.
To create visual coherency, AR applications require an accurate and robust environment \emph{lighting estimation}, which ensures that virtual objects blend naturally with the physical environment.

\begin{figure}[t]
    \centering
\begin{subfigure}[b]{0.485\columnwidth}
        \centering
        \includegraphics[width=\linewidth]{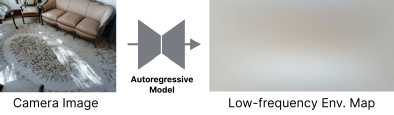}
        \caption{Autoregressive Lighting Estimation}
        \label{subfig:background_autoregressive_lighting_estimation}
    \end{subfigure}\quad
    \begin{subfigure}[b]{0.485\columnwidth}
        \centering
        \includegraphics[width=\linewidth]{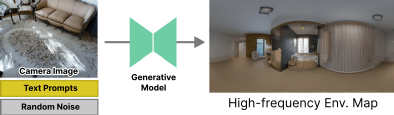}
        \caption{Generative Lighting Estimation}
        \label{subfig:background_generative_lighting_estimation}
    \end{subfigure}
    \vspace{-.5em}
    \caption{
        Comparison between autoregressive and generative lighting estimation methods.
        \textnormal{(\ref{subfig:background_autoregressive_lighting_estimation}) An autoregressive lighting estimation system, Xihe~\cite{xihe_mobisys2021}, estimates omnidirectional low-frequency lighting information from camera images with autoregressive models. The low-frequency lighting information misses important visual details, as visualized in the example environment map.  (\ref{subfig:background_generative_lighting_estimation}) Generative lighting estimation models can create high-frequency environment lighting estimation results from limited environment observations. The estimation process can be conditioned in several ways, such as partial environment observations and text prompts.
}
    }
    \vspace{-.5em}
    \label{fig:background}
\end{figure}

To address this challenging task, traditional systems often employ autoregressive models~\cite{xihe_mobisys2021} to extract parametric lighting information from AR device camera images (shown in Fig.~\ref{subfig:background_autoregressive_lighting_estimation}). While these methods effectively capture coarse environmental lighting conditions, they struggle to reliably capture high-frequency details, such as geometric features of the environment. To address the critical needs of high-frequency information in environment lighting, recent works~\cite{somanath-envmapnet,yang2023diffusion,wang2022stylelight} have sought new solutions by leveraging the advancements in \emph{controllable generative models}. The primary reason is that these models have the potential to generate fine-grained environmental details (shown in Fig.~\ref{subfig:background_autoregressive_lighting_estimation}), thereby empowering AR systems with a more realistic rendering effect. However, it is challenging to achieve robust mobile AR lighting estimation with generative models in real-world environments. We identify two key challenges below.

First, robust lighting estimation demands accurate and timely estimations.
Therefore, it is crucial to ensure that the system can provide reliable and accurate estimation outputs under diverse and complex lighting conditions.
Additionally, generative models often exhibit long inference latencies, which slow down the estimation process.
Directly integrating generative model-based lighting estimation methods may not satisfy the needs of real-time mobile AR experiences.
Second, training and evaluating robust lighting estimation models require careful attention to data bias.
Specifically, the datasets used must be curated to reflect diverse and balanced lighting conditions to ensure fair and generalizable performance.
However, our measurement study on existing datasets reveals distribution biases in key lighting properties, such as intensity and color temperature, which in turn affect the generalization and robustness of several recent lighting estimation models~\cite{wang2022stylelight,akimoto2022diverse}.

In this paper, we address the above issues with \sysname, a novel generative lighting estimation system for mobile AR.
To support robust lighting estimation, \sysname builds on two key ideas: \1 AR context-guided generative estimation and \2 edge-device collaborative estimation. It leverages \emph{multimodal AR context} to guide 360$^\circ$ HDR environment map estimation using generative diffusion models.
To ensure temporal consistency, \sysname uses a multi-output strategy to generate environment maps that best match real-time lighting conditions.
To improve estimation quality and responsiveness, \sysname applies color appearance matching for fine-grained adjustments to environment maps.

At the core of \sysname, the lighting estimation process in \sysname revolves around a \emph{two-step generative pipeline} that estimates 360$^\circ$ HDR environment maps from \emph{partial LDR environment observations}.
Our key design insight is to separate the generative model training objective into two domains: \emph{LDR environment map completion} and \emph{high-intensity pixel value estimation}.
This novel learning objective design addresses the practical challenge of the scarcity of high-quality lighting estimation training data, allowing us to leverage pre-trained large models to tackle each generation step effectively.
Beyond its two-step generative pipeline, \sysname leverages AR context---\emph{semantic maps} to guide visual details and \emph{ambient light} sensor data to inform lighting intensity and color temperature.
This context-guided generative estimation design effectively helps \sysname to tackle the challenge of estimating omnidirectional environment maps from limited environment observations on mobile AR devices.

To evaluate \sysname, we conduct a comprehensive evaluation that includes in-lab deployment tests, data-driven evaluations, and a user study.
In the deployment test, \sysname excels in supporting virtual object rendering quality compared to three representative baselines: unwrapping a mirror ball (physical reference)~\cite{debevec2006image}, ARKit (commercial)~\cite{arkit}, and LitAR (academic)~\cite{zhao2022litar}.
Quantitatively, we compare \sysname with state-of-the-art lighting estimation models~\cite{Phongthawee2023DiffusionLight,wang2022stylelight,akimoto2022diverse,Gardner2017} and show that \sysname outperforms the best performing baseline, DiffusionLight~\cite{Phongthawee2023DiffusionLight}, by up to 56\%.
Our user study also confirms the effectiveness and robustness of \sysname, showing at least a 12\% improvement in quality ratings compared to StyleLight, the second-best method.
Furthermore, we introduce a robustness testing protocol, which leverages our augmented Laval dataset to test the estimation accuracy under diverse lighting intensity and temperature conditions.
We observe that \sysname consistently achieves low estimation errors across diverse lighting conditions, whereas the estimation quality of baseline methods varies significantly with changes in lighting.

Related works on mobile AR lighting estimation systems seek to extract environment information from physical light probes~\cite{prakash2019gleam}, user dynamics, and learning-based solutions~\cite{xihe_mobisys2021,gardner2019deep,wang2022stylelight,yang2023diffusion}.
While physical light probes provide the most comprehensive environment observations, their use is limited by the need for physical setup.
Consequently, AR applications often use learned models to estimate lighting from camera images.
Over the past couple of decades, learning-based methods have evolved from discovering scene lighting cues from image details, such as highlights and shadows~\cite{yu1999inverse}, to regressing omnidirectional environment lighting representations~\cite{zhao2020pointar,xihe_mobisys2021,gardner2019deep}.
However, autoregressive models cannot effectively tackle the environment information generation in lighting estimation.
For example, Xihe~\cite{xihe_mobisys2021} provides real-time low-frequency lighting estimation for AR applications but lacks support for detailed environment reflections in object rendering.
In contrast, recent advances in generative models enable the estimation of highly detailed environment maps~\cite{wang2022stylelight,yang2023diffusion}, opening new possibilities for visually coherent rendering.
Our work explores the integration of image-generative models into AR systems to enable high-quality environment lighting estimation, with novel AR context-aware design to improve generation speed, accuracy, and robustness.

We summarize our main contribution as follows:

\begin{itemize}[leftmargin=.12in,topsep=4pt]
    \item We design and implement \sysname, a generative lighting estimation system that allows mobile AR applications to acquire environment lighting to support visually coherent AR experiences. The relevant research artifacts are at \url{https://github.com/cake-lab/CleAR}.
    \item We introduce a two-step generative pipeline to produce more accurate and visually coherent environment maps aligned with the user’s physical surroundings. This design effectively overcomes the limited environmental sensing of AR devices and smartly leverages multi-modal AR context to achieve accurate estimation.
    \item To further improve estimation accuracy and real-time robustness, we develop a set of refinement techniques that work in tadam with the generative estimation pipeline. These refinement components use a multi-output estimation strategy to address ambiguity and adapt the estimated environment color based on real-time AR device camera observations.
    \item We evaluate \sysname through both quantitative and qualitative experiments, comparing it against SoTA lighting estimation systems and models on a commonly used dataset~\cite{Gardner2017} and a robustness-focused variant generated by us.
    A user study based on perceptual quality ratings shows that \sysname outperforms all baselines, scoring 12\% higher than the second-best method, with 7\% lower rating variance.
\end{itemize}

\section{Background}
\label{sec:background}

\para{Environment Lighting Representations.}
Lighting estimation has been a long-standing research question in the vision and graphics community~\cite{li2023spatiotemporally}.
Pioneering works have established several methods for capturing high-fidelity lighting of physical environments.
For instance, environmental lighting can be captured with mirror balls, panoramic cameras, and photometric stereo techniques~\cite{debevec2006image,ward2008high,vlasic2009dynamic}.
The captured environment lighting can also be represented in various formats, including parametric format~\cite{green2003spherical}, neural representations~\cite{pandey2021total}, and image-based format~\cite{debevec2006image}.
In modern computer graphics, mirror ball-based capturing and image-based lighting are widely used for accurate lighting rendering. However, mirror ball setups can be cumbersome in practice~\cite{prakash2019gleam}, making image-based lighting a more practical approach for mobile AR, which this work adopts.
Image-based lighting relies on acquiring high-dynamic range (HDR) environment maps that capture omnidirectional lighting at a specific location. HDR images are critical because they preserve the full range of luminance---including bright highlights and subtle shadows---often lost in standard low-dynamic range (LDR) images~\cite{dufaux2016high}. To enable realistic lighting effects and photorealistic virtual content, this work focuses on generating HDR environment maps.

\para{Lighting Estimation in Mobile AR.}
In mobile AR, accurately estimating environment lighting is crucial for creating immersive visual experiences when overlaying virtual content on physical environments.
Such application includes virtual try-ons~\cite{zhao2023multi} and interactive games~\cite{scorpio2022calibration}.
Obtaining accurate environmental lighting in mobile AR faces several additional challenges compared to traditional image— or video-based lighting estimation tasks.
For example, mobile AR devices typically have limited environmental sensing capabilities, such as the field of view (FoV) of their cameras.
Because visually coherent AR experiences require omnidirectional environment lighting, the limitations in environment sensing render lighting estimation in mobile AR a highly ambiguous process and, thus, an ill-posed problem.
Traditional methods~\cite{Song2019,Gardner2017,gardner2019deep} often aim to solve this problem using deep models, which result in lengthy computation times.
However, mobile AR applications require timely environment lighting updates to maintain visual coherence in temporally changing lighting environments.
Recently, mobile system researchers have begun leveraging enhanced device capabilities~\cite{xihe_mobisys2021} and user-in-the-loop interactions~\cite{zhao2022litar} to acquire more reliable environmental information, thereby improving the accuracy and robustness of lighting estimation in mobile AR.
Despite these advances, current state-of-the-art methods still fall short of providing the accuracy and robustness required for reliable lighting estimation in mobile AR applications.

\para{Conditional Generative Models.}
Recent years have witnessed significant advancements in the data synthesis capabilities of generative models~\cite{yang2023diffusion,croitoru2023diffusion}.
Various model architectures, such as Generative Adversarial Networks (GANs)~\cite{goodfellow2020generative}, Variational Autoencoders (VAEs)~\cite{kingma2013auto}, have achieved notable success in generating realistic images, text, and other forms of data.
However, the content generation process in these models typically cannot be conditioned. In other words, users of these models cannot control the outputs through external guidance or specific input signals.
Conditional generative models extend generative models by incorporating additional input information, referred to as conditions, to guide the generation process.
These conditions can range from class labels in image generation~\cite{mirza2014conditional} to textual descriptions in text-to-image models~\cite{reed2016generative} or even semantic features for structured data generation~\cite{hong2018conditional}.
In this work, we base \sysname on the ControlNet~\cite{zhang2023adding} architecture, a type of state-of-the-art conditional diffusion model.
ControlNet works by extending a pre-trained diffusion model, such as Stable Diffusion~\cite{rombach2021highresolution}, with additional neural network layers that allow external conditions to influence the diffusion process.
Specifically, ControlNet uses a trainable copy of the diffusion model's encoder to encode the input conditioning latent code and merge it with the diffusion model's latent code.
By doing so, it maintains the strong generative priors of the base diffusion model while allowing the generation process to be guided by external conditioning inputs.

\section{Motivation and Challenges}
\label{sec:preliminary}

\begin{figure}[t]
    \centering
\begin{subfigure}[b]{0.425\columnwidth}
        \centering
        \includegraphics[width=\linewidth]{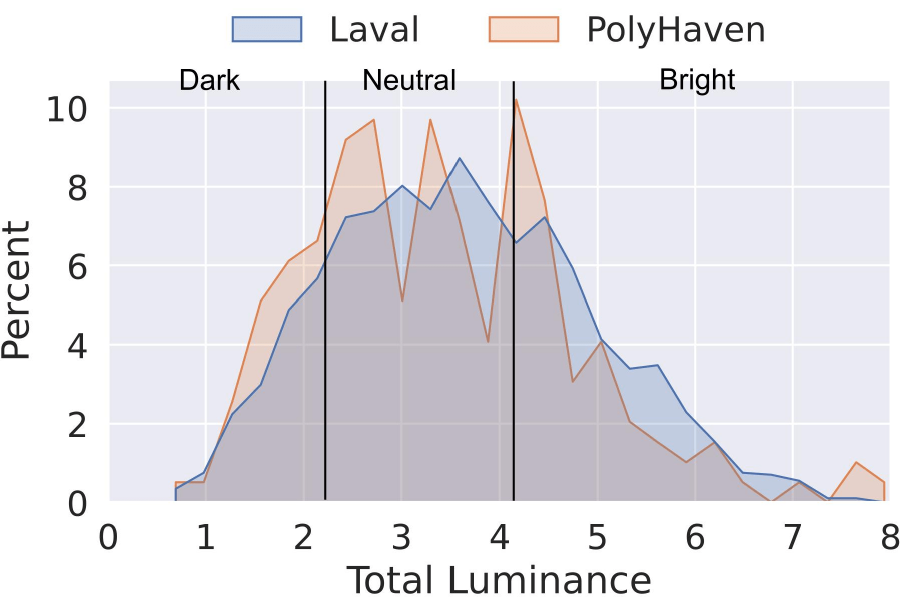}
        \caption{Light Intensity}
        \label{subfig:color_intensity}
    \end{subfigure}\quad
    \begin{subfigure}[b]{0.425\columnwidth}
        \centering
        \includegraphics[width=\linewidth]{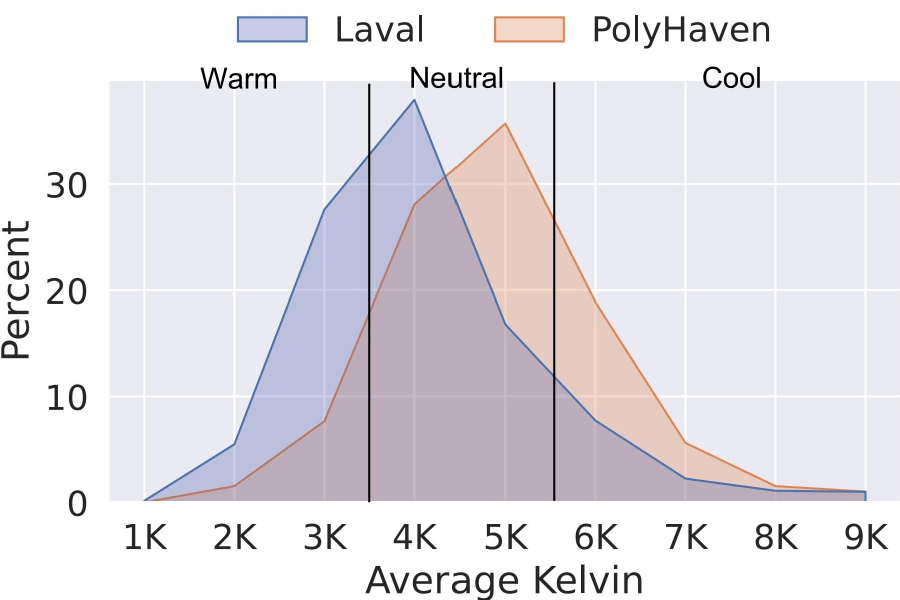}
        \caption{Color Temperature}
        \label{subfig:color_temperature}
    \end{subfigure}
\caption{
Distributions of lighting condition in two datasets.
        \textnormal{We analyze the distributions of light intensity and color temperature on environment maps collected from the Laval indoor dataset and PolyHaven.
Both datasets exhibit data biases.
        For example, a significant part of the PolyHaven data items lies in the cool color range. Also, both datasets have an imbalanced distribution between neutral and other lighting conditions.}
    }
    \vspace{-1em}
    \label{fig:preliminary_data_dist}
\end{figure}

Our analysis uses two standard lighting estimation datasets: the \emph{Laval dataset}~\cite{Gardner2017}, an academic open research dataset, and the \emph{PolyHaven} website~\cite{polyhaven}, a royalty-free HDR environment map data website.
On these data, we calculate the \emph{light intensity} as the total luminance of a given HDR environment map image, along with the environment map image's color temperature values in Kelvin.
Additional details of analysis setup, measurement calculations, and technical details are included in Appendix~\S\ref{sec:additional_lighting_property_measurement_details}.
Figure~\ref{fig:preliminary_data_dist} shows the lighting condition distributions with annotated human-perceivable categories for light intensity and color temperature.

Our analysis reveals two critical challenges. First, \emph{real-world lighting conditions are complex and diverse}.
Both datasets include a wide range of light intensity and color temperature values.
In particular, the measured light intensity spans from as low as 0.6 to over 8 in luminance values, capturing both dim and brightly lit scenes. Similarly, the color temperature varies widely, ranging from below 1000K, typical of incandescent lighting, to over 9000K, which is seen in artificial lights. This suggests that real-world lighting conditions have complex and diverse properties.
In many real-world scenarios, environmental lighting conditions often vary across scenes and over time.
For example, indoor rooms lit by sunlight often exhibit neutral to warm color temperatures, while those illuminated by LED sources tend to appear cooler.
Therefore, lighting estimation systems must have rich built-in knowledge of diverse environmental lighting conditions, as well as adaptive application-time policies to react to real-time environment lighting changes.
To address this challenge, in \S\ref{sec:sys_design}, we introduce an end-to-end system that takes multimodal \emph{AR context} to guide the generation of a visually coherent HDR environment map with color refinement and color palette matching.

Another key challenge revealed by our study is the \emph{inherent data biases}, which pose challenges to training and evaluating generative models in lighting estimation systems.
These biases in training data can lead to overfitting on dominant lighting conditions and poor generalization to underrepresented lighting features. For instance, models trained predominantly on neutral or cool lighting conditions may fail to produce plausible estimations under warm or high-intensity illumination. Furthermore, biased data distributions in testing data can skew evaluation results, making it difficult to evaluate the robustness of lighting estimation systems.
To address the data bias issue,
We introduce a \emph{data balancing technique} in Appendix~\S\ref{subsec:dataset_generation}, and present a robustness testing protocol in \S\ref{sec:evaluation}, through which we evaluate several recent lighting estimation models.

\section{\sysname Design}
\label{sec:sys_design}

\begin{figure*}[t]
\centering
    \includegraphics[width=.9\linewidth]{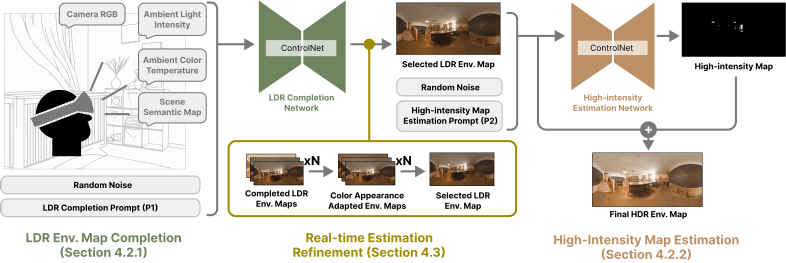}
    \caption{Overview of \sysname.
        \textnormal{
            \sysname uses four types of AR context data to guide a two-step high-quality lighting generation pipeline. The first step (\S\ref{subsubsec:ldr_environment_map_completion}) completes partial environment observations in the LDR pixel range domain. The completed environment maps will be post-processed and selected using our real-time estimation refinement components (\S\ref{subsec:real_time_estimation_refinement}) and then passed on to the subsequent high-intensity map estimation step (\S\ref{subsubsec:high_intensity_map_estimation}).
            Finally, \sysname outputs an HDR environment map by combining the completed LDR environment map and the high-intensity map.
}
}
\label{fig:sys_overview}
\end{figure*}

\subsection{Overview}
\label{subsec:design_overview}

At a high level, estimating HDR omnidirectional lighting from low-FoV LDR environment observations is a process of significant information increase.
To tackle this challenging problem, \sysname leverages the recent advancements in conditional image generative models~\cite{zhang2023adding} to create missing environment lighting information, guided by AR context data.
The key design of \sysname is a novel integration of image generative models to mobile AR lighting estimation systems to ensure high-quality and robust estimation results.

As illustrated in Figure~\ref{fig:sys_overview}, the design of \sysname revolves around an AR context-guided generative lighting estimation pipeline (\S\ref{subsec:context_guided_lighting_estimation}).
\sysname estimates omnidirectional environment lighting, a 360$^\circ$ HDR environment map image, from multimodal AR context data. In this work, we consider four types of AR data, including low-FoV LDR images, ambient light intensity and color temperature, and scene semantic map.
Our context-guided generation design addresses the key limitation of image generative models.
While generative models excel at image generation in general, producing accurate HDR environment maps requires precise conditioning to ensure visual consistency with real-world scenes, especially under complex lighting conditions and limited environmental observations.
We address this challenge by \emph{using AR context data to guide generation}, offering both coarse- and fine-grained conditioning.

Another critical aspect of robust mobile AR lighting estimation is the timely update of estimation results to address changes in environmental lighting.
However, the high inference latency of generative models limits the frequency of lighting updates.
To address this challenge, \sysname employs an estimation refinement workflow (\S\ref{subsec:real_time_estimation_refinement}) that adapts the color profile of estimated environment maps to real-time AR camera images, ensuring temporally consistent lighting estimation.
Specifically, \sysname first employs a multi-output generation strategy (\S\ref{subsubsec:estimation_output_selection}) to create accurate yet diverse estimation results for unobserved environments.
Then, \sysname uses several techniques (\S\ref{subsubsec:color_apperance_adaptation} and \S\ref{subsubsec:estimation_output_selection}) to refine the previously estimated environment maps and select the best one as the estimation output.

\vspace{-0.5em}
\subsection{Context-Guided Generative Lighting Estimation}
\label{subsec:context_guided_lighting_estimation}

\subsubsection{LDR Environment Map Completion.}
\label{subsubsec:ldr_environment_map_completion}

In the first step, \sysname completes the environment map from limited camera observations, treating the standard 8-bit pixel range as the LDR range.
Our LDR completion model builds on the ControlNet~\cite{zhang2023adding} architecture with pre-trained Stable Diffusion~\cite{rombach2021highresolution} model weights.
We choose ControlNet for its flexible controllability using text and image conditioning, which integrates well with the AR context.
The pre-trained StableDiffusion model provides strong prior knowledge learned from large-scale image generation datasets\footnote{Note that \sysname's design is not tied to ControlNet nor Stable Diffusion; other conditional image generative models can serve as drop-in replacements to improve the performance of \sysname in the future.}.
These priors enable us to train environment map generation models without starting from scratch entirely, also allowing for better generalization to diverse real-world scenarios.
While Stable Diffusion excels at general image generation, it must be fine-tuned on environment map data to adapt it to our task of generating panoramas in the correct format and style.

During the completion process, we encode four types of AR context data into three input modalities for ControlNet: \1 a camera RGB image as direct environmental observation, \2 a scene semantic map for structural conditioning, and \3 a text prompt encoding ambient light intensity and color temperature.
As part of input preparation, we stitch multi-view camera RGBs into 360 $^\circ$ panoramic environment map images in advance.
This process can be extended with more sophisticated reconstruction methods like~\cite{zhao2022litar} or combined with multi-user observation sharing to increase the observed environments.
Similarly, environment semantic maps are also processed into 360 $^\circ$ panoramic images, following the ADE20K~\cite{zhou2017scene} definition with 150 instance labels.

To encode ambient lighting conditions, we convert numerical sensor data into natural language descriptions of the environment's lighting characteristics.
We use the following text prompt template to describe ambient lighting conditions.
The lighting condition words are marked in bold font.

\vspace{.5em} \begin{center}
  \begin{minipage}{\linewidth}
    \textbf{P1}: \textit{A panoramic photo of an indoor room. The room is in a \textbf{[dark/neutral/bright]} lighting condition. The room has a \textbf{[warm/neutral/cool]} ambient color.}
  \end{minipage}
  \label{prompt:ldr_completion}
\end{center}

The ambient light property labels are created based on partial LDR image pixels with empirically derived threshold values. The light intensity label is defined by the mean pixel intensity~\cite{bolduc2023beyond} with values between 0.25 and 0.40 as \textit{neutral}, and the rest of the values as \textit{dark} and \textit{bright}.
We define the color temperature label based on the mean pixel color temperature: values between 3500K and 5500K are considered \textit{neutral}, while values outside this range are categorized as \textit{warm} or \textit{cool}~\cite{noguchi1999effect}.

\subsubsection{High-intensity Map Estimation.}
\label{subsubsec:high_intensity_map_estimation}

\begin{figure}[t]
\centering
    \includegraphics[width=.8\linewidth]{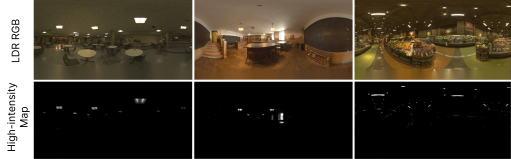}
\caption{
        Visualization of high-intensity pixels.
        \textnormal{We decompose three examples of HDR environment maps into an LDR RGB component (row 1) and a high-intensity component (row 2). The high-intensity pixels, i.e., bright light spots on the environment map, are extracted using Equation~\eqref{eq:hdr_scaling}.}
    }
\label{fig:hdr_decomposition}
\end{figure}

The second step of our pipeline is to estimate high-intensity components in the completed LDR environment maps.
The high-intensity components describe environment light intensities and directionalities, which are critical for rendering photorealistic highlight and shadow effects on AR objects.
However, due to limited camera capabilities, mobile AR devices often fail to capture high-intensity lighting information.
As shown in Figure~\ref{fig:hdr_decomposition}, we decompose each HDR environment map into two parts using Equation~\eqref{eq:hdr_scaling}---an LDR representation and a high-intensity component (in the form of a high-intensity map)---for visualization.
We define a \textit{high-intensity map} as a 360$^\circ$ LDR image that describes the high-intensity light positions, directions, and pixel values from the original HDR environment map.

\vspace{-.7em}
\begin{equation}
    I_m = {2.0} / ({1 + e^{- I_i}}) - 1.0
    \label{eq:hdr_scaling}
\end{equation}

To generate high-intensity maps, we again condition a generation model on the completed LDR environment map produced in \S\ref{subsubsec:ldr_environment_map_completion}.
One of the key challenges in building reliable high-intensity map estimation models is the scarcity of high-quality HDR environment maps.
For example, the Laval dataset includes only about 2K data items.
Training generative models on small datasets can be highly unreliable.
Additionally, pre-trained backbone models, like Stable Diffusion~\cite{rombach2021highresolution}, cannot be directly used for high-intensity map estimation because they are trained on LDR images.

To address these challenges, we propose a novel learning formulation to allow high-intensity estimation in the LDR pixel range domain.
Specifically, our design includes two key steps.
First, we use a scaling transformation (Equation\eqref{eq:hdr_scaling}) to convert HDR environment maps into LDR and high-intensity maps.
Then, we fine-tune a pre-trained generative model, using the transformed Laval dataset, to learn the mapping from LDR environment maps to high-intensity maps.
This is feasible because \emph{both input and output are represented in the LDR pixel range}.
More importantly, the high-intensity values often align with simple LDR image features, such as bright spots and light strips, making high-intensity map estimation a simpler task compared to the LDR completion task.
During training and inference, we use the following text prompt to guide the high-intensity estimation process in producing consistent image style and reducing visual artifacts.

\vspace{.5em} \begin{center}
  \begin{minipage}{\linewidth}
        \textbf{P2}: \textit{A grayscale panoramic image describing the bright spots of an indoor room. Brighter spots get more bright color. Regions without light sources should stay pure black.}
  \end{minipage}
  \label{prompt:ldr_completion}
\end{center}

This novel design allows us to overcome the HDR data availability challenge and use the small amount of ground-truth HDR data from the Laval dataset to train the high-intensity estimation model.
Combined with the completed LDR environment map from the first step (\S\ref{subsubsec:ldr_environment_map_completion}), our generative lighting estimation pipeline outputs a 360$^\circ$ HDR environment map.

\subsection{Real-time Estimation Refinement}
\label{subsec:real_time_estimation_refinement}

\begin{figure}[t]
\centering
    \includegraphics[width=0.9\linewidth]{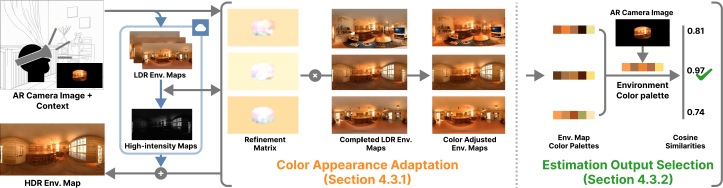}
\caption{
        Estimation refinement workflow.
        \textnormal{\sysname uses a hybrid architecture to offload heavy generative model inference to the edge server while adaptively refining estimation results on AR devices in real time. We adopt a multi-output generation strategy to tackle the estimation ambiguity. On the client, our system first matches the color appearances between the completed environment maps and the real-time camera observations (\S\ref{subsubsec:color_apperance_adaptation}). This operation helps improve estimation accuracy in challenging lighting conditions and enables real-time adaptation of the generative estimation result. Then, our system client chooses the best estimation result based on real-time environment observation. The chosen environment map is combined with its corresponding high-intensity map estimation to create the final HDR environment map.}
    }
\label{fig:generation_control}
\end{figure}

Completing 360$^\circ$ environment maps from low-FoV environment observations is an inherently ambiguous process.
Although AR context data helps condition the generation process, each generation can take several seconds---leading to potential mismatches between the current AR camera image and the one used during generation.
To address this, we introduce a multi-output estimation process co-designed with our two-step generation pipeline. Our key insight is to generate multiple completion variants and select the one that best matches the real-time AR view. At a high level, the process consists of three steps as illustrated in Figure~\ref{fig:generation_control}.

Our multi-output generation design not only enables \sysname to output accurate lighting estimation results, but also provides practical support for responsive lighting estimation in the real-time AR application workflow.
The high memory footprint of generative models, often exceeding 2–4 GB during inference~\cite{rombach2021highresolution}, makes them unable to run on resource-constrained mobile platforms that must concurrently handle rendering and sensing tasks. In contrast, our on-device refinement method enables the estimated environment maps to adapt to minor changes in environmental lighting, thereby reducing the need for frequent inference of the generative model.
To facilitate real-time deployments, \sysname adopts an edge-device collaborative estimation architecture, similar to recent works~\cite{ben2022edge,xihe_mobisys2021}. Specifically, LDR environment map completion and high-intensity map estimation can be offloaded to an edge server, while lightweight color refinement and result selection algorithms are executed on-device.

\subsubsection{Color Appearance Adaptation}
\label{subsubsec:color_apperance_adaptation}
Addressing unpredictable changes in environmental lighting is crucial for supporting temporally consistent rendering of AR objects.
Therefore, \sysname employs a lightweight on-device refinement technique to refine the multi-output LDR completion results to adapt their color appearances to real-time AR camera images.
While other additional metrics like semantic consistency or structural similarity are also important for environment map quality, we consider recovering these properties as orthogonal problems that can be addressed separately in real-time environment reconstruction systems~\cite{theodorou2022visual}.

Our design selects the AR camera image as the adaptation target because the ultimate goal of mobile AR lighting estimation is to support the visually coherent insertion of AR objects into camera images.
Additionally, our on-device real-time color appearance adaptation design is well-positioned to address the real-time estimation needs in mobile AR.
By applying color adaptation to the estimated environment maps, \sysname achieves consistently high estimation accuracy with few generative model inference requests.

A recent work, DiffusionLight~\cite{Phongthawee2023DiffusionLight}, uses a similar strategy that has to invoke large quantities of inferences to achieve desirable quality results.
However, generative models, particularly diffusion-based models, typically suffer from long inference latency.
Under the default setting, DiffusionLight requires 60 diffusion model inference calls to generate one estimation result.
As we will show in \S\ref{sec:evaluation}, DiffusionLight incurs a very high end-to-end inference time (359.9s), making its multi-output estimation strategy infeasible for real-time AR applications.
In comparison, using our color adaptation technique, \sysname can achieve better estimation accuracy than DiffusionLight while using approximately 110X less time.
Additional implementation details and testing examples of color refinement are included in Appendix \S\ref{sec:on_device_real_time_refinement_details_appendix}.

\subsubsection{Estimation Output Selection.}
\label{subsubsec:estimation_output_selection}

As the final step of the real-time estimation refinement process, \sysname selects the best environment map image from the previously refined LDR environment map images.
In this process, \sysname searches for the best estimation candidate by identifying the highest cosine similarity between the color palettes of the observation and estimation environment map images.
The color palettes are selected using the K-means algorithm, which identifies the five most common colors for each environment map image.
The total cosine similarity is calculated as the sum of per-color cosine similarity between color palettes.
The selected LDR environment map will later be combined with its corresponding high-intensity map to form a complete HDR environment map for mobile AR applications.

\section{Implementations}
\label{sec:implementation}

\begin{figure*}[t]
\centering
\resizebox{0.9\linewidth}{!}{\begin{minipage}{\linewidth}
        \begin{subfigure}[b]{0.32\linewidth}
            \centering
            \includegraphics[width=\linewidth]{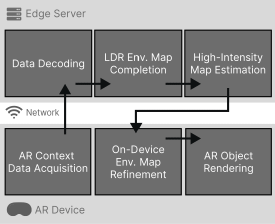}
            \caption{System Architecture Overview}
            \label{subfig:implementation_arch}
        \end{subfigure}\hfill
        \begin{subfigure}[b]{0.32\linewidth}
            \centering
            \includegraphics[width=\linewidth]{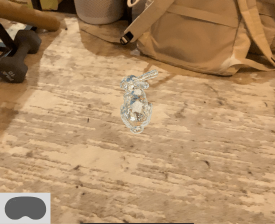}
            \caption{Example AR Application}
            \label{subfig:implementation_device}
        \end{subfigure}\hfill
        \begin{subfigure}[b]{0.32\linewidth}
            \centering
            \includegraphics[width=\linewidth]{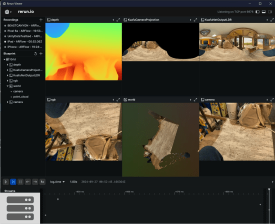}
            \caption{Edge Server Data Visualization}
            \label{subfig:implementation_edge}
        \end{subfigure}
    \end{minipage}} \caption{System implementation and AR application integration. Figure~\ref{subfig:implementation_arch} shows an overview of the software components of \sysname. Figure~\ref{subfig:implementation_device} depicts an AR application screenshot of a rendered virtual bunny with \sysname estimated environment lighting. Figure~\ref{subfig:implementation_edge} illustrates the visualization component at the edge server for the collected AR data.}
\label{fig:implementation}
\end{figure*}

\subsection{System Implementation}
\label{subsec:system_implementation}

We implement \sysname based on a recent AR data streaming framework, ARFlow~\cite{zhao2024arflow}, which enables low-latency bidirectional data streaming between the AR device and the edge server.
We also implement an example object placement AR application with \sysname to demonstrate the end-to-end rendering results.
Figure~\ref{fig:implementation} shows an overview of the \sysname architecture.
We implement the client as a Unity3D\footnote{Unity3D: \url{https://unity.com}} package.
The client is responsible for streaming and receiving data from mobile AR devices and providing AR applications with estimated environment maps.
We use ARFlow's built-in data collection feature to capture camera RGB, depth, and device tracking data provided by ARFoundation\footnote{ARFoundation: \url{https://docs.unity3d.com/Packages/com.unity.xr.arfoundation@6.0/}}, Unity's low-level AR framework.
The captured data is streamed to our server via the ARFlow gRPC service.
The server component is a Python-based service for the generative lighting estimation pipeline.
During generative model inference, we combine the RGB and other context information using the \texttt{MultiControlNet} pipeline from the HuggingFace \texttt{Diffusers}\footnote{Diffusers: \url{https://huggingface.co/docs/diffusers}} library to run different ControlNet models.
We configure the ControlNet model inference to use the UniPC~\cite{zhao2024unipc} sampler with 20 steps.
Our ControlNet models output environment maps at a resolution of \texttt{512x256}.
\texttt{xformers}\footnote{xFormers: \url{https://facebookresearch.github.io/xformers/}} is used to accelerate model inference.

\subsection{Training and Testing Data Generation}
\label{subsec:training_and_testing_data_generation}

We curate a large-scale dataset for training and testing \sysname. For LDR completion model training, we assemble 30K LDR environment map images from two prominent LDR indoor environment map sources: the Matterport3D dataset~\cite{Matterport3D} and Structured3D datasets~\cite{zheng2020structured3d}. Matterport3D dataset provides real-world captured environment maps, and Structured3D dataset consists of photorealistic synthetic ones.
For high-intensity map estimation model training, we use the training split of the Laval dataset~\cite{Gardner2017} to create 1,489 pairs of LDR environment map images and corresponding high-intensity map images.
To evaluate \sysname under diverse lighting conditions, we create variants of the Laval test set that preserve the original environment map visual context. We generate these variants by uniformly scaling the environment map images.
For light-intensity editing, all color channels are scaled equally; for color temperature editing, only the red and blue channels are adjusted~\cite{afifi2020deep}.
See Appendix~\S\ref{subsec:dataset_generation} for additional details on the data generation process, and Appendix~\S\ref{subsubsec:appendix_model_training} for details of model training.

\section{Evaluation}
\label{sec:evaluation}

To comprehensively evaluate \sysname, we use a combination of quantitative and qualitative methods to assess key aspects of system performance, including generation latency, output quality, and user perception.
We conduct a lab-based evaluation using an example AR application built on top of \sysname to assess rendering quality and latency.
Our data-driven evaluation compares \sysname to several state-of-the-art lighting estimation methods~\cite{Phongthawee2023DiffusionLight,wang2022stylelight,akimoto2022diverse} using both the standard three-sphere protocol and our proposed robustness testing protocol under varied lighting conditions.
In addition, we use an online survey to understand the impact of lighting estimations on human perceptual preferences.
Our results demonstrate that \sysname achieves high-quality lighting estimation with low latency.
For example, our end-to-end estimation results show that \sysname achieves the best performance in two image-based metrics while taking significantly less time.
Participants also rated \sysname-generated lighting as producing better renderings for most virtual objects.

\subsection{Testbed-Based Evaluation}
\label{subsec:testbed_based_evaluation}

\subsubsection{Testbed Setup.}
We conduct our evaluation using an example AR application running on a 4th-generation 11-inch iPad Pro with a built-in LiDAR sensor. The edge server is a high-end workstation with an Intel Core i9-13900K CPU, 128GB of RAM, and an NVIDIA RTX 4090 GPU. The edge server runs an Ubuntu 22.04 operating system. The mobile device and edge server are connected via an 802.11ac Wi-Fi network. To ensure consistent and reliable results, all system runtime measurements are performed 10 times, and the reported values are the average across these runs.

\begin{figure*}[t]
\centering
\includegraphics[width=.9\linewidth]{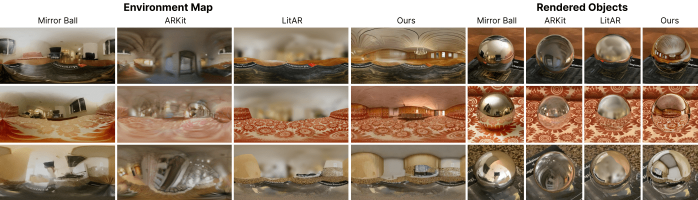}
    \caption{AR object rendering results. We show AR virtual object rendering qualitative comparison to other environment lighting acquired by unwrapping a physical mirror ball~\cite{debevec2006image} (reference), ARKit~\cite{arkit} (commercial), and LitAR~\cite{zhao2022litar} (academic). Compared to ARKit, \sysname estimates the environment map with significantly more realistic visual details. Compared to LitAR, \sysname can generate more visually coherent far-field environment map details, while LitAR can only generate blurry ones.
    }
    \label{fig:eval_qualitative_testbed}
\end{figure*}

\begin{table}[t]
\centering
\caption{System performance breakdown. (Left) We show the breakdown of system latency under the default generative estimation settings, which use five LDR completion outputs and do not use semantic maps in the input. In the left table, on-device operations are marked with \raisebox{0pt}[0pt][0pt]{\colorbox{gray!20}{\texttt{D}}}, edge operations are marked with \raisebox{0pt}[0pt][0pt]{\colorbox{gray!20}{\texttt{E}}}, and network operations are marked with \raisebox{0pt}[0pt][0pt]{\colorbox{gray!20}{\texttt{N}}}. (Right) We show the end-to-end system execution time (excluding network time) under various system configurations. Our system can generate an environment map as fast as 2,383 ms. Under the default system configuration, which is marked in \textbf{bold} font, the end-to-end execution time is 3,269 ms.
}
\vspace{-1em}
\label{tab:eval_latency_breakdown}
\begin{minipage}[t]{0.48\linewidth}
\vspace{0pt}
\centering
\small
\begin{tabularx}{\linewidth}{@{}Xr@{}}
\toprule
\textbf{System Component}   & \multicolumn{1}{c}{\textbf{Avg. Time (ms)}} \\ \midrule
\raisebox{0pt}[0pt][0pt]{\colorbox{gray!20}{\texttt{D}}} Device data preparation     &      0.5 ($\pm$ 0.1)      \\
\raisebox{0pt}[0pt][0pt]{\colorbox{gray!20}{\texttt{N}}} AR Data Offloading          &      31 ($\pm$ 10.1)      \\
\raisebox{0pt}[0pt][0pt]{\colorbox{gray!20}{\texttt{E}}} LDR completion x5              &      1957 ($\pm$ 9.3)      \\
\raisebox{0pt}[0pt][0pt]{\colorbox{gray!20}{\texttt{E}}} LDR env. map retrieval       &       72 ($\pm$ 8.1)      \\
\raisebox{0pt}[0pt][0pt]{\colorbox{gray!20}{\texttt{D}}} Estimation refinement           &     3 ($\pm$ 0.2)      \\
\raisebox{0pt}[0pt][0pt]{\colorbox{gray!20}{\texttt{E}}} Refinement result synchronization              &     10 ($\pm$ 0.6)   \\
\raisebox{0pt}[0pt][0pt]{\colorbox{gray!20}{\texttt{N}}} High-intensity map estimation              &     1166 ($\pm$ 10.7)   \\
\raisebox{0pt}[0pt][0pt]{\colorbox{gray!20}{\texttt{N}}} High-intensity map retrieval       &       30 ($\pm$ 8.9)      \\
\bottomrule
\end{tabularx}
\end{minipage}
\hfill
\begin{minipage}[t]{0.48\linewidth}
\vspace{0pt}
\centering
\small
\begin{tabularx}{\linewidth}{@{}Xr@{}}
\toprule
\textbf{System Configuration}  & \multicolumn{1}{c}{\textbf{Avg. Time (ms)}} \\ \midrule
LDR completion x1 (RGB)    &            2383 ($\pm$ 11.4)                \\
LDR completion x1 (RGB + Semantics)   &            3021 ($\pm$ 51.3)                \\
LDR completion x3 (RGB)   &           2825 ($\pm$ 17.6)                 \\
LDR completion x3 (RGB + Semantics)   &           4526 ($\pm$ 31.5)                 \\
\textbf{LDR completion x5 (RGB)} &             \textbf{3269 ($\pm$ 19.5)}                 \\
LDR completion x5 (RGB + Semantics) &             5855 ($\pm$ 38.5)                 \\
LDR completion x7 (RGB) &             3741  ($\pm$ 14.2)               \\
LDR completion x7 (RGB + Semantics) &             8082  ($\pm$ 56.9)               \\
\bottomrule
\end{tabularx}
\end{minipage}
\end{table}

\subsubsection{Qualitative Visual Comparison.}
We showcase \sysname's lighting estimation quality with our testbed AR application in real-world environments.
We followed the experiment setup of LitAR~\cite{zhao2022litar}, a recent lighting estimation mobile system, to compare the quality of lighting estimation on rendered virtual sphere images.
In Figure~\ref{fig:eval_qualitative_testbed}, we show the visual comparison between two baselines: LitAR~\cite{zhao2022litar} and ARKit~\cite{arkit}. We also include a physical mirror ball as a reference for physical environment lighting.
For fair comparison, we use the same near-field reconstruction data from LitAR as the RGB input to \sysname.
Overall, we observe that \sysname outputs the best environment map image quality among estimated environment maps (ARKit and LitAR).
\sysname generates environment maps with more visual details than LitAR and fewer artifacts than ARKit.
Particularly, \sysname can generate detailed far-field environments while LitAR can only generate blurry ones.
When comparing rendered virtual mirror balls to the physical reference mirror ball, \sysname produces the closest visual match.

\subsubsection{System Performance Breakdown}
Table~\ref{tab:eval_latency_breakdown} shows \sysname's component-wise latency and its end-to-end execution latency under different configurations.
In the default configuration, \sysname takes RGB images and ambient light data from the AR device as input to generate five LDR completion variants and one corresponding high-intensity map.
The end-to-end execution takes 3,269 ms, primarily consisting of the generative LDR completion time.
When combined with semantics in LDR completion, the end-to-end execution takes 5,855 ms.
We observe that the main impact factor for the LDR completion process is the number of environment map outputs.
Increasing generation output introduces a near-linear yet slow increase in the estimation latency.
Although our default system configuration uses five outputs, the estimation latency is only about double that of using a single estimation output.
Note that these long execution time could be masked by pipelining per-frame requests to a powerful edge server, if needed.
However, with our on-device refinement, we have the feasibility to skip expensive generative inferences yet still meet the real-time and quality goals.
Specifically, once the high-intensity map is generated, \sysname can generate visually coherent HDR environment maps in real-time by matching the LDR environment map to the current camera view's color appearance. If the environmental lighting conditions do not change much, our design allows for the reuse of expensive generative results across frames without sacrificing the visual quality.
To put \sysname's latency in context, we note that a recent AR-specific lighting estimation system  LitAR~\cite{zhao2022litar} needs 46.6/134.38ms to generate low/high quality near-field portion of the environment map.
Another AR-specific system Xihe can achieve real-time lighting estimation (20.1ms) but can only produce low-fidelity spherical harmonics coefficients. In contrast, \sysname can deliver real-time HDR environment maps, which are much higher quality lighting information.

\subsection{Data-Driven Evaluation}
\label{subsec:experiment_setups}

\begin{figure}[t]
\centering
    \vspace{2em}
    \includegraphics[width=0.9\linewidth]{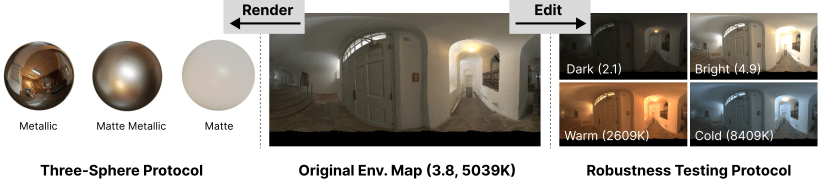}
\caption{
        Visualizations of two experiment protocols for data-driven evaluation.
        \textnormal{For the three-sphere protocol (left), we use environment maps (middle) from the Laval dataset to render virtual spheres of different material properties.
For the robustness testing protocol (right), we use the robustness testing dataset, consisting of environment maps from Laval with edited ambient light intensity and color temperature.
        As an example, we show an environment map (middle) with four variants with different light intensities and color temperatures. The results (right) of light intensity augmentation are marked as \emph{dark} and \emph{bright}, and the results of color temperature augmentation are marked as \emph{warm} and \emph{cold}.}
    }
    \label{fig:primer_data_aug}
\end{figure}

\subsubsection{Experiment Setups.}
We quantitatively evaluate the lighting estimation quality of \sysname against three state-of-the-art lighting estimation models: DiffusionLight~\cite{Phongthawee2023DiffusionLight}, StyleLight~\cite{wang2022stylelight}, and OmniDreamer~\cite{akimoto2022diverse}.
Among these baselines, DiffusionLight is most similar to \sysname because it uses both diffusion models and a multi-output generation pipeline.
We use the Laval test set and its augmented variants (\S\ref{subsec:training_and_testing_data_generation}) to evaluate the accuracy and robustness of lighting estimation methods using the following two protocols:

\begin{itemize}
\item
\para{The three-sphere evaluation protocol}~\cite{wang2022stylelight,Phongthawee2023DiffusionLight} uses a proxy method to evaluate the HDR lighting estimation accuracy through pixel-wise error measurements on three rendered spheres with different material properties.
Specifically, the used sphere materials are: \emph{matte}, \emph{silver matte}, and \emph{silver}.
Figure~\ref{fig:primer_data_aug} shows a set of rendered virtual spheres used in this evaluation protocol.
Each material represents a mixture of specific reflectance and roughness properties.
Compared to direct environment map-wise comparison, this evaluation protocol helps us understand the \emph{impact of lighting estimation on object rendering quality}.

\item
\para{The robustness testing protocol} is designed by us to test the \emph{generalizability and robustness of lighting estimation systems} under different lighting conditions, specifically changes in light intensity and color temperature.
In this protocol, we test lighting estimation systems on the Laval variants and compare estimated environment maps with the ground truth to calculate pixel-wise errors.
The comparison is performed in the LDR image domain to capture lighting condition-related color appearance differences between environment map images.
\end{itemize}

\para{Evaluation Metrics.}
Following prior works~\cite{wang2022stylelight,akimoto2022diverse,Phongthawee2023DiffusionLight,gardner2019deep}, we use the FID score~\cite{Seitzer2020FID} to measure the diversity of generated environment maps.
For evaluating the accuracy of environment maps, we follow ~\cite{zhao2022litar} to use RMSE.
We use three image-based metrics---scale-invariant Root Mean Square Error (si-RMSE)~\cite{eigen2014depth}, Angular Error~\cite{legendre:2019:deeplight}, and RMSE---to evaluate lighting estimation accuracy on rendered virtual spheres.
The si-RMSE and RMSE quantify pixel-wise differences, with si-RMSE being insensitive to global intensity scaling. Angular Error focuses on differences in pixel chromaticity, emphasizing the assessment of environment lighting color properties.
We follow ~\cite{Phongthawee2023DiffusionLight} to map the 0.1st and 99.9th value percentiles to 0 and 1 when calculating RMSE.

\subsubsection{End-to-End Visual Quality and Performance.}
\begin{figure}[t]
\centering
    \includegraphics[width=.9\linewidth]{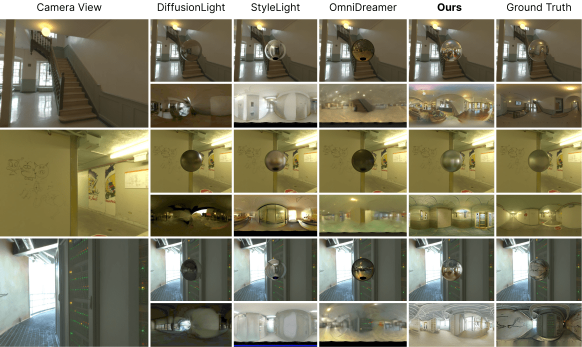}
\caption{
        Qualitative comparisons.
        \textnormal{We show examples of lighting estimation results on three scenes and the corresponding rendered virtual sphere images.
For each method, we show rendered virtual spheres (row top) and estimated environment maps (row bottom).
        Note, OmniDreamer can only output LDR environment maps.
        }
    }
    \label{fig:eval_qualitative_comparisions}
\end{figure}

\begin{figure}[t]
\centering
    \centering

    \begin{minipage}{0.7\linewidth}
        \centering
        \includegraphics[width=\linewidth]{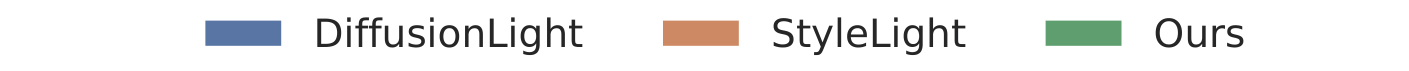}
    \end{minipage}

\vspace{0.5em}

    \begin{subfigure}[b]{0.25\columnwidth}
        \centering
        \includegraphics[width=\linewidth]{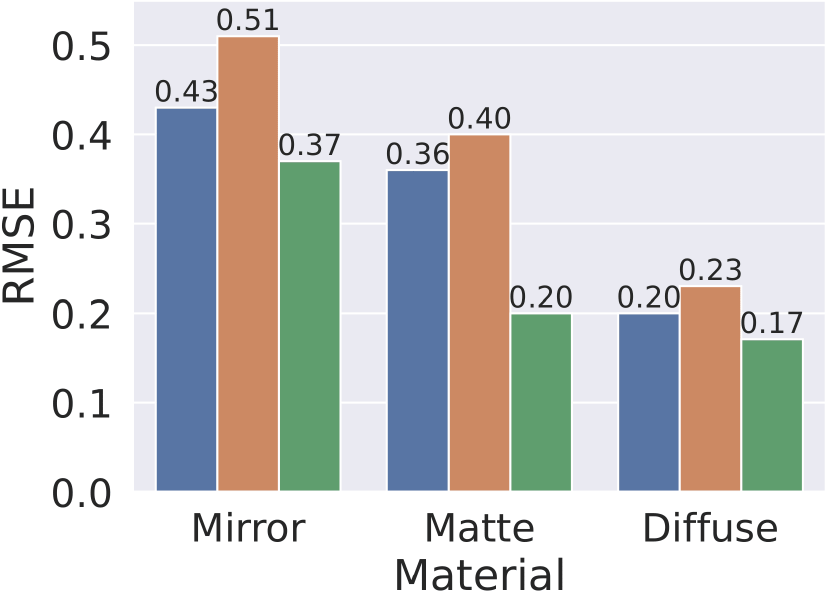}
        \caption{RMSE}
        \label{subfig:e2e_three_sphere_rmse}
    \end{subfigure}\quad
    \begin{subfigure}[b]{0.25\columnwidth}
        \centering
        \includegraphics[width=\linewidth]{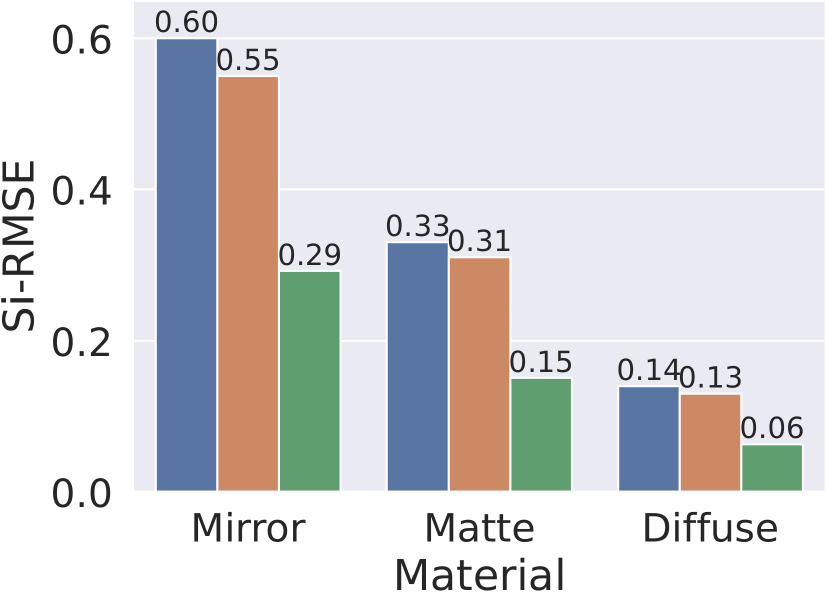}
        \caption{Si-RMSE}
        \label{subfig:e2e_three_sphere_sirmse}
    \end{subfigure}\quad
    \begin{subfigure}[b]{0.25\columnwidth}
        \centering
        \includegraphics[width=\linewidth]{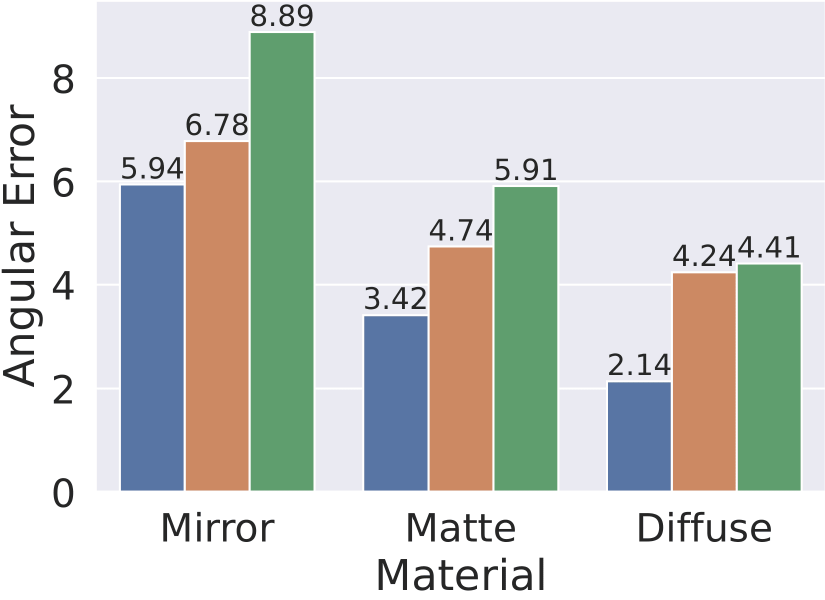}
        \caption{Angular Error}
        \label{subfig:e2e_three_sphere_angular_error}
    \end{subfigure}\quad
    \begin{subfigure}[b]{0.14\columnwidth}
        \centering
        \includegraphics[width=\linewidth]{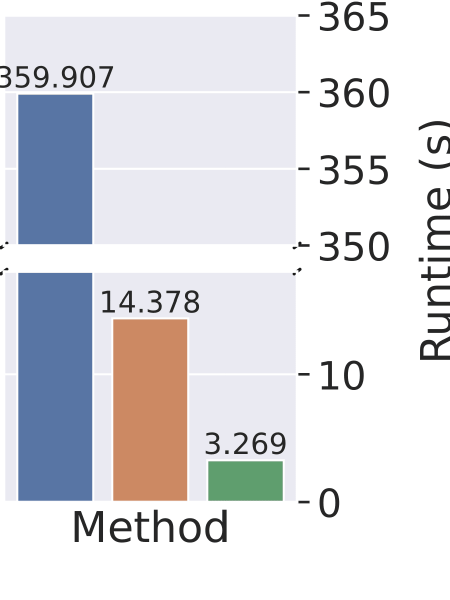}
        \caption{Latency}
        \label{subfig:e2e_latency}
    \end{subfigure}

    \vspace{-1em}

    \caption{
        End-to-end qualitative comparison.
        \textnormal{We compare the end-to-end estimation accuracy and latency between \sysname and two state-of-the-art works~\cite{wang2022stylelight,Phongthawee2023DiffusionLight}. We test estimation accuracy through the three-sphere evaluation protocol, which directly measures the visual quality impact of environment lighting on AR virtual object rendering.
        \sysname achieves the lowest si-RMSE and RMSE values, on all three testing object material types in the three-sphere evaluation protocol. However, \sysname has slightly higher angular errors on the mirror and matte spheres.
        Moreover, \sysname achieves this competitive accuracy with significantly lower end-to-end latency compared to the other generative model-based methods.
}
    }
    \label{fig:eval_e2e_quantitative}
\end{figure}

In Figure~\ref{fig:eval_qualitative_comparisions}, we show qualitative comparisons between \sysname and baseline methods in three scenes, representing the neutral, warm, and cool lighting conditions.
Compared to other methods, \sysname allows more accurate overall color tones rendering and creates more diverse reflection details on the rendered virtual spheres.
Figure~\ref{fig:eval_e2e_quantitative} presents a quantitative comparison between \sysname and two SoTA methods~\cite{Phongthawee2023DiffusionLight,wang2022stylelight} that can output HDR environment maps in both rendering quality and estimation latency.
Under the same environmental observations, \sysname achieves the lowest scale-invariant and normalized RMSE values across all three virtual sphere types.
Noticeably, for the mirror sphere, the most challenging material of the three, \sysname achieves approximately $50\%$ reduction in the scale-invariant RMSE values.
On the other hand, we observe that \sysname generates environment maps with slightly higher angular errors.
Upon further inspection, we suspect that the diverse environment map image details potentially caused the higher angular error because the angular error metric is sensitive to pixel color differences.
Furthermore, \sysname achieves the lowest lighting estimation latency by large margins.
In particular, compared to DiffusionLight~\cite{Phongthawee2023DiffusionLight}, a diffusion model-based method, \sysname takes significantly less time (110X).

\para{Impact of Context Data Availability.}
We follow the setups in~\cite{wang2022stylelight,Phongthawee2023DiffusionLight} to use a single centered view of a 75$^\circ$ horizontal FoV camera as the default configuration for \sysname.
This configuration represents a baseline performance of \sysname because our system can take in more AR context data during runtime.
Table~\ref{tab:eval_three_sphere} shows \sysname's performance under different configurations of camera FoVs, the number of input views, and the usage of semantic maps.
Note that the ambient light data has been included in all tests in Table~\ref{tab:eval_three_sphere} because it is inseparable from our estimation pipeline design.
The added environment observations through new views and increased FoVs can both improve the accuracy of lighting estimation.
Additionally, full-scene semantic maps are important guiding information for high-accuracy lighting estimation.
Full-scene semantic maps can often be obtained as floor maps or scene design layouts in real-world applications.
This ablation study suggests that sharing scene semantics can greatly benefit lighting estimation accuracy with our system.

\begin{table}[t]
\centering
\caption{
Impact of context data.
We evaluate \sysname's estimation accuracy under different configurations based on the three-sphere protocol.
The first row shows the default configuration of \sysname, marked in \raisebox{0pt}[0pt][0pt]{\colorbox{gray!20}{gray}}, which already outperforms DiffusionLight and StyleLight in both si-RMSE and RMSE.
As more context data is used---larger FoV, environmental semantics, and more camera observations, \sysname can further improve the rendering quality.
}
\label{tab:eval_three_sphere}
\vspace{-1em}
\resizebox{\textwidth}{!}{
\begin{tabular}{cccccccccc}
\toprule
    & \multicolumn{3}{c}{\textbf{Scale-invariant RMSE $\downarrow$}}       & \multicolumn{3}{c}{\textbf{Angular Error $\downarrow$}}           & \multicolumn{3}{c}{\textbf{RMSE $\downarrow$}}         \\
\multirow{-2}{*}{\textbf{Configuration}}                     & Diffuse                       & Matte& Mirror                        & Diffuse                      & Matte                        & Mirror                       & Diffuse                      & Matte                        & Mirror                       \\ \hline
\rowcolor{gray!10}
\multicolumn{1}{l}{RGB 75$^\circ$ FoV 1 view}          & 0.06   & 0.15   & 0.29  & 4.42 & 5.92 & 8.89 & 0.17 & 0.21 & 0.37 \\
\multicolumn{1}{l}{RGB+semantics 75$^\circ$ FoV 1 view}          & 0.06   & 0.14   & 0.28  & 4.41 & 5.91 & 8.89 & 0.17 & 0.20 & 0.37 \\
\multicolumn{1}{l}{RGB+semantics 110$^\circ$ FoV 1 view}         & 0.05   & 0.13   & 0.27  & 4.19 & 5.77 & 8.66 & 0.16 & 0.19 & 0.35 \\
\multicolumn{1}{l}{RGB+semantics 110$^\circ$ FoV 3 views}        & 0.03   & 0.10   & 0.24  & 3.89 & 5.21 & 8.23 & 0.13 & 0.17 & 0.31 \\
\multicolumn{1}{l}{RGB+semantics 110$^\circ$ FoV 5 views}        &  \underline{0.02}  &  \underline{0.09}  & \underline{0.22}  & \underline{3.71} & \underline{5.13} & \underline{8.17} & \underline{0.11} & \underline{0.16} & \underline{0.29} \\
\multicolumn{1}{l}{RGB+full semantics 110$^\circ$ FoV 3 views}   & \textbf{0.02}   & \textbf{0.07}   & \textbf{0.20}  & \textbf{3.59} & \textbf{4.96} & \textbf{8.09} & \textbf{0.10} & \textbf{0.15} & \textbf{0.27} \\ \bottomrule
\end{tabular}
}

\end{table}

\begin{figure}[t]
    \centering
\includegraphics[width=0.6\linewidth]{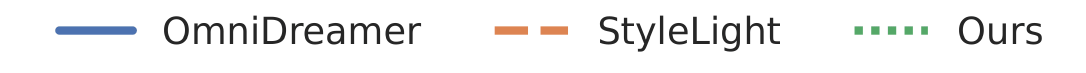}

    \begin{subfigure}[b]{0.375\columnwidth}
        \centering
        \includegraphics[width=\linewidth]{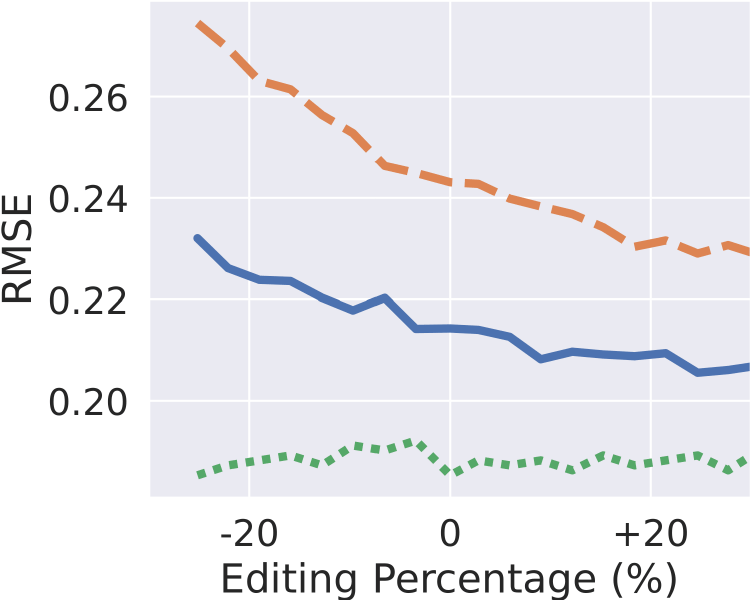}
        \caption{Light Intensity}
        \label{subfig:intensity_benchmark}
    \end{subfigure}\quad
    \begin{subfigure}[b]{0.375\columnwidth}
        \centering
        \includegraphics[width=\linewidth]{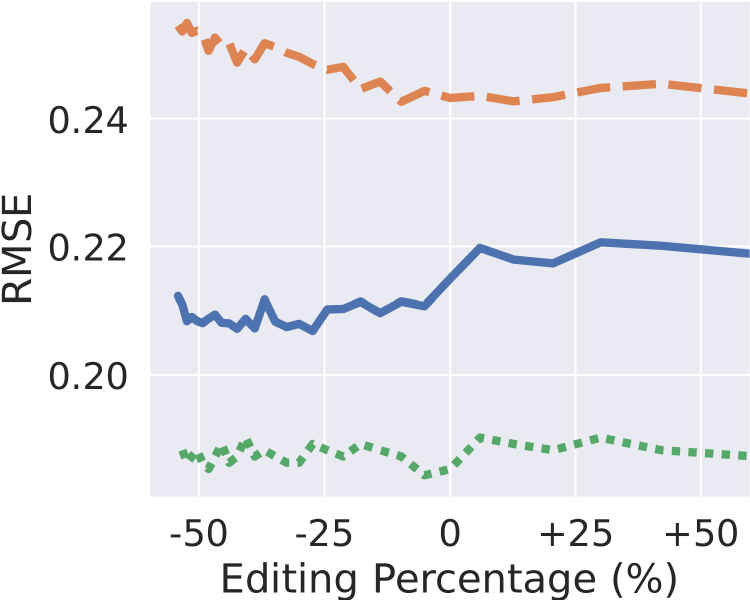}
        \caption{Color Temperature}
        \label{subfig:temperature_benchmark}
    \end{subfigure}
    \vspace{-.7em}
    \caption{
        Robustness testing result.
        \textnormal{We test the robustness of our lighting estimation results against two recent lighting estimation models~\cite{akimoto2022diverse,wang2022stylelight} using our robustness testing protocol.
        \sysname shows consistently lower estimation error rates on different lighting conditions. This observation indicates \sysname can generalize better and provide more consistent result quality in different lighting conditions.}
    }
    \label{fig:robustness_testing}
\end{figure}

\para{Estimation Robustness Evaluation.}
We test the lighting estimation robustness of \sysname under several environmental lighting conditions with the robustness testing protocol.
We compare \sysname against two recent lighting estimation methods that are built on top of generative models: OmniDreamer~\cite{akimoto2022diverse} and StyleLight~\cite{wang2022stylelight} \footnote{We were unable to complete the robustness testing for DiffusionLight~\cite{Phongthawee2023DiffusionLight} within a reasonable time given its high inference cost.}.
Figure~\ref{fig:robustness_testing} shows the RMSE values calculated on the estimated LDR environment maps.
We choose to compare the LDR environment maps, instead of the HDR environment maps, because lighting conditions mainly affect the accuracy of LDR environment map estimation.
We observe that the accuracy of both OmniDreamer and StyleLight is affected by the changing lighting intensities and color temperatures.
While \sysname, on the other hand, shows consistently lower estimation error rates under different lighting conditions.
Particularly, these two models exhibit much higher lighting estimation errors on lower environmental lighting intensity and warmer color temperature.
The observed error increases potentially are caused by the unbalanced training datasets used by these models.

\subsubsection{System Ablation Study.}
In this section, we evaluate the estimation accuracy of \sysname by comparing its performance across different system configurations and design choices.

\begin{figure}[t]
\centering
\begin{subfigure}[b]{0.5\linewidth}
        \centering
        \resizebox{\linewidth}{!}{\begin{tabular}{@{}ccc@{}}
        \toprule
        \multirow{2}{*}{\textbf{Method}} & \textbf{FID $\downarrow$} & \textbf{FID $\downarrow$} \\
                                         & (full env. map)           & (selected regions)        \\ \midrule
        Gardner et al.~\cite{Gardner2017}              & 307.5                     & 197.4                     \\
        OmniDreamer~\cite{akimoto2022diverse}          & \underline{106.3}         & \underline{46.2}          \\
        StyleLight~\cite{wang2022stylelight}           & 137.7                     & 97.2                      \\
        DiffusionLight~\cite{Phongthawee2023DiffusionLight} & 207.2                     & 193.5                     \\
        Ours                             & \textbf{86.3}             & \textbf{44.31}            \\ \bottomrule
        \end{tabular}
        } \vspace{1.1em}
        \caption{LDR Completion Output Diversity}
        \label{tab:eval_ldr_fid}
    \end{subfigure}\quad
\begin{subfigure}[b]{0.2\linewidth}
        \centering
        \includegraphics[width=\linewidth]{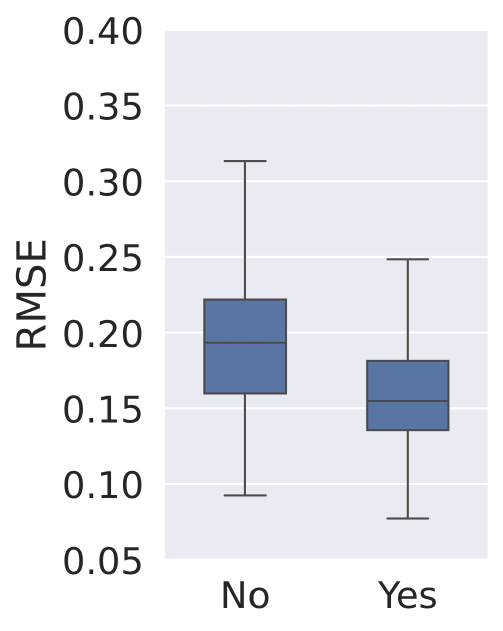}
        \caption{Color Refinement}
        \label{subfig:color_matching}
    \end{subfigure}\begin{subfigure}[b]{0.187\linewidth}
        \centering
        \includegraphics[width=\linewidth]{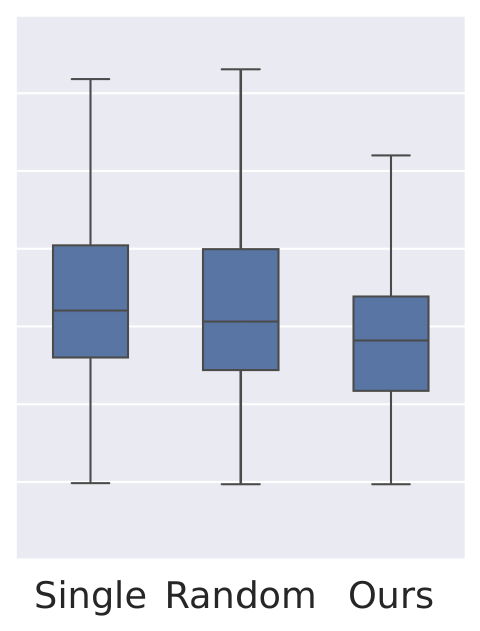}
        \caption{Output Selection}
        \label{subfig:output_selection}
    \end{subfigure}
    \vspace{-.5em}
    \caption{
        Evaluation of generative estimation refinement.
        \textnormal{
        (\ref{tab:eval_ldr_fid}) Our method achieves the lowest FID scores, showing strong generation diversity.
        (\ref{subfig:color_matching}) Our color appearance matching technique improves estimation accuracy.
        (\ref{subfig:output_selection}) Our output selection method reduces RMSE and variance compared to baselines.}
    }
    \label{fig:eval_generation_control}
\end{figure}

\para{Analysis of LDR Environment Map Completion.}
For the completion of the LDR environment map, we specifically examine the pixel-wise errors and the overall diversity of the environment map content.
The former assesses \sysname's LDR lighting estimation accuracy, and the latter evaluates the generation richness of \sysname generative estimation pipeline.
For \sysname, we use the LDR environment maps after applying the color appearance adaptation.
Following~\cite{akimoto2022diverse,wang2022stylelight}, we adopt two methods for measuring the generation content diversity: \1 calculate the FID score on the full estimated LDR environment map, and \2 converting the environment map into a cube map and calculate FID scores on each face \emph{without the top and bottom faces} as these two faces containing little information.
In Table~\ref{tab:eval_ldr_fid}, we show that our LDR environment map completion model can output environment map images with greater diversity than state-of-the-art models.
Our LDR completion model outperforms other models by 27\% to 370\% in the first calculation method and 4\% to 344\% in the second calculation method.
This observation confirms the effectiveness of our large LDR environment map completion dataset.
Next, we compare the accuracy of the LDR environment map completion.
Under the same environment observations as~\cite{wang2022stylelight,Phongthawee2023DiffusionLight}, our method achieves comparable results to DiffusionLight even though our LDR completion model is based on a smaller-sized pre-trained diffusion model.

\para{Analysis of Estimation Refinement.}
We evaluate the performance of our on-device estimation refinement components on the impacts of lighting estimation accuracy.
In Figure~\ref{subfig:color_matching}, we show that our color appearance matching technique improves the overall estimation accuracy by 31\%.
The estimation refinement technique allows our system to achieve high-quality estimation with limited estimation outputs.
By default, our system only requires five generation outputs while DiffusionLight~\cite{yang2023diffusion} requires more than 90 generation outputs.
Reducing the required generation output will also reduce the estimation latency and computation resources requirements during deployment.
Next, we evaluate the effectiveness of our generation output selection policy.
In Figure~\ref{subfig:output_selection}, we show that our color-matching technique can reduce the estimation error rate compared to other selection methods.
We observe that, with our output selection policy and a total of five generation outputs, \sysname can reduce 15\% of the estimation error compared to using a single generation output.
Furthermore, when using five-generation outputs, our generation output selection component can reduce the average estimation error rate by 11\% compared to random selection.

\subsection{Perceptual User Study}
\label{sec:user_study}

We conduct an online user study, via Qualtrics\footnote{Qualtrics: \url{https://www.qualtrics.com/}}, to assess the impact of lighting estimation on perceived rendering quality.
Our study was approved by our organization's Institutional Review Board (IRB) and then distributed through personal and professional networks.

\subsubsection{Study Protocol}
\label{subsec:quality_assesment_sutdy}

\begin{figure}[t]
\centering
    \centering
\includegraphics[width=.85\linewidth]{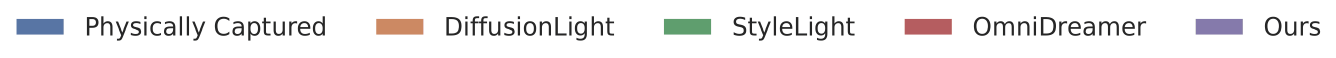}
    \includegraphics[width=.9\linewidth]{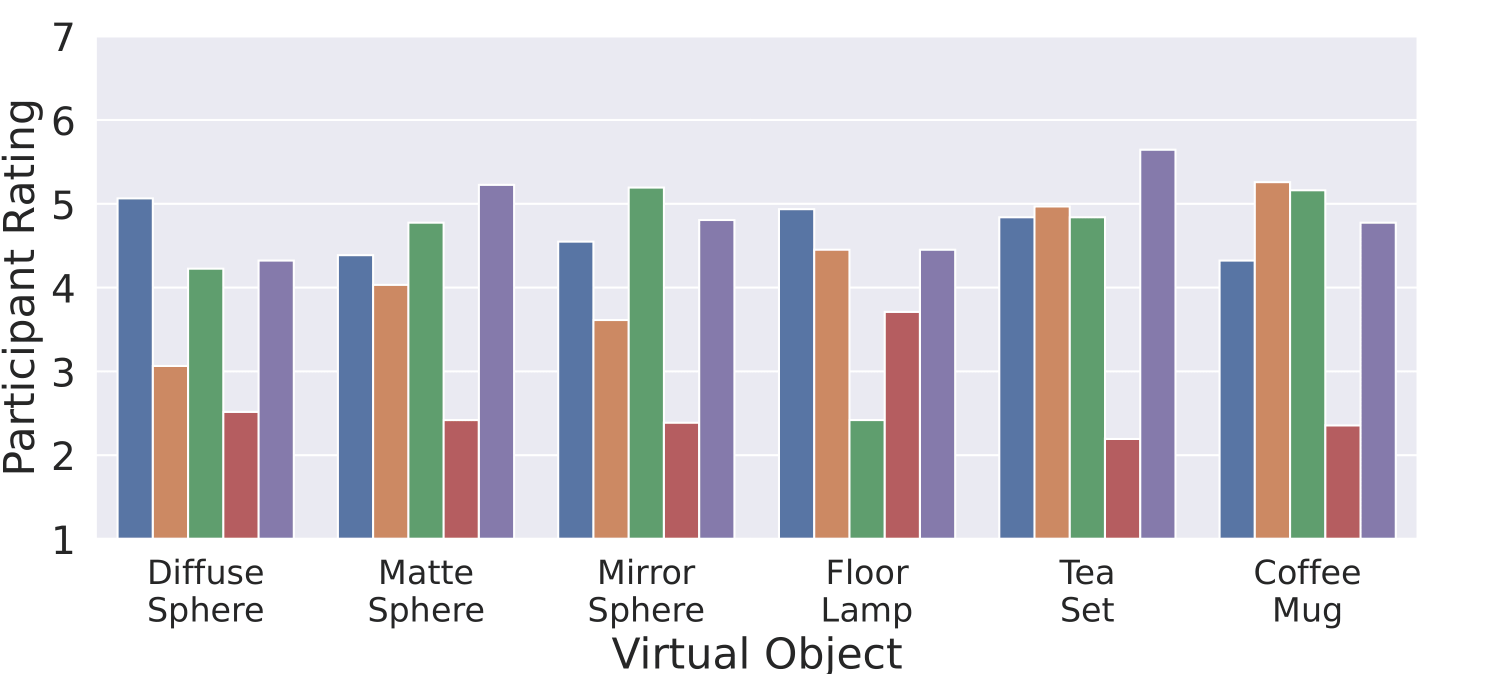}
    \caption{
Human perceptual preference comparisons.
        \textnormal{We compare virtual object rendering quality between \sysname and four other methods. Based on responses from 31 participants, \sysname received the highest average rating of 4.87, outperforming the second-best method, StyleLight, by 12\%. \sysname also shows more consistent quality, with a lower standard deviation.
}
    }
    \label{fig:result_user_study}
\end{figure}

Our survey consists of three chapters.
The first chapter surveys the participants' past experiences with mobile AR and their impressions of virtual object rendering qualities with existing mobile AR applications.
The second chapter is a training section that guides participants in completing the quality assessment study.
This chapter first shows a quality rating question where the participant will be given five images of the same virtual sphere rendered under lighting conditions from five different sources.
Participants are instructed to rate the visual qualities using a Likert scale of 1 to 7, where 1 represents the lowest and 7 represents the highest quality.
To avoid bias, participants are not informed which lighting source was used for each image.
An example rating for this question is shown to the user for training purposes.
Next, we will show a follow-up question to ask participants for feedback, specifically on the quality issues of images generated by our system.
The last chapter is the formal quality assessment, consisting of six question groups.
Each group uses different environments and virtual object setups, representing environments with varying lighting conditions and objects with various geometries and materials.
Six virtual objects are used in total, including a diffuse sphere, a matte sphere, a mirror sphere, a floor lamp, a tea set, and a coffee mug.
Specifically, lighting conditions used in these questions were generated using \sysname, DiffusionLight~\cite{Phongthawee2023DiffusionLight}, StyleLight~\cite{wang2022stylelight}, OmniDreamer~\cite{akimoto2022diverse}, and physically captured HDR environment maps.
Each group follows the same format used in the training chapter.
The lighting conditions used to generate the images are randomized in order and not revealed to participants.

\subsubsection{Results and Analysis.}
\label{subsec:effectiveness_of_ar_context}

We received responses from 31 participants (48\% Respondents are familiar with multiple areas of graphics technology, most commonly image editing (51.6\%), followed by video editing (41.9\%) and 3D modeling and rendering (35.5\%).
25.8\% reported no familiarity with specific graphics technologies.
58\% had used AR devices, most commonly mobile phones or tablets (48\%), followed by Meta Quest (29\%).

Figure~\ref{fig:result_user_study} shows the visual quality assessment results from the last chapter.
Excluding ratings for objects rendered with physically captured lighting, participants rated \sysname-generated lighting as producing better renderings for 4 out of the six virtual objects.
The visual quality rating of \sysname ranked the top with an average score of 4.87 across all responses.
This score is about 12\% better than that of the second-best method, StyleLight, which received an average rating of 4.35.
Additionally, \sysname has a lower overall standard deviation in ratings (1.56) compared to StyleLight (1.67), indicating more robust estimation quality.

Interestingly, in two question groups---the tea set and coffee mug---participants rated \sysname’s renderings as higher quality than those produced using physically captured environment maps.
Upon further inspection, we found that, while physically captured environment lighting more accurately reflects the radiance conditions in the scene, human perception tends to favor the slightly brighter outputs generated by \sysname.
We hope this finding encourages future research to design and evaluate lighting estimation systems with considerations of human perceptual preferences.

\section{Related Work}
\label{sec:related_work}

\para{Environment Lighting and Rendering.}
Pioneering works have established several ways of capturing high-fidelity physical environment lighting and representing it in digital formats.
Omnidirectional HDR environment map is a commonly adopted solution~\cite{debevec2006image} for representing the environment lighting because it stores the incoming radiance information that can be easily integrated with modern computer graphics rendering.
Environment maps can be commonly captured using mirror balls, 360$^\circ$ cameras, or bracketed image stitching~\cite{debevec2023recovering,brown2007automatic}.
A recent work, GLEAM~\cite{prakash2019gleam}, incorporates the mirror ball-assisted environment lighting capturing process into AR applications.
However, the requirement for the presence of the mirror ball limits the practicality of AR device usage.
Instead, our system uses the generative estimation approach to provide flexible, high-quality lighting estimation.
Additionally, HDR environment maps can be challenging to obtain on AR devices due to many sensor limitations.
Recent lighting estimation systems~\cite{xihe_mobisys2021,zhao2022litar} can only output LDR ones.
But our novel two-step generative estimation design allows \sysname to support HDR lighting estimation.

\para{Image Generative Models.}
Recently, generative models have demonstrated impressive capabilities in image synthesis, as well as generating audio, 3D models, and other data modalities~\cite{yang2023diffusion,croitoru2023diffusion}.
The generative diffusion model~\cite{ho2020denoising} has attracted significant attention from the research community because of its high-quality image generation capabilities.
The generative diffusion model uses a physical diffusion process inspired by the Gaussian signal denoising generation process.
Combining this novel generation process with the large model parameter sizes, generative diffusion models outperform generative models with prior architectures, such as GAN~\cite{goodfellow2020generative} or VAE~\cite{esser2021taming}.
Several recent works~\cite{chen2022text2light,akimoto2022diverse,wang2022stylelight,Phongthawee2023DiffusionLight} also propose to adopt generative models of different architectures to solve the generic lighting estimation problem. Most notably, DiffusionLight~\cite{Phongthawee2023DiffusionLight} achieves state-of-the-art estimation accuracy with diffusion models.
In this work, we present a generative model-based approach to lighting estimation tailored for mobile AR applications, addressing key challenges in achieving visually coherent and fast generation.

\para{Context-Aware Mobile System.}
Context-aware computing is a classical computing paradigm in which applications sense and adapt to contextual information~\cite{chen2000survey}.
Early research in context-aware AR systems~\cite{starner1998visual} demonstrated that important environment information can be extracted from camera frames for task planning and decision-making.
In recent years, new developments of AR systems have also sought to leverage broad types of environment context information to assist several AR tasks and achieve better user experiences~\cite{lam2021a2w,tahara2020retargetable,wang2020capturar}.
Our work leverages four types of AR context data---camera RGB, scene semantic map, and ambient light intensity and color temperature---to guide environment map estimation and align virtual rendering with real-world lighting.

\section{Discussion}
\label{sec:concluding_remarks}

\para{Application Use Cases.}
\sysname's design enables a wide range of mobile AR applications that demand visually coherent environment lighting. For instance, \sysname can significantly enhance virtual object rendering in entertainment applications by ensuring that rendered objects consistently blend with real-world scenes under diverse lighting conditions. Additionally, \sysname is also well suited for commercial applications. For example, in interior retail applications, \sysname enables the creation of visually appealing virtual furniture that accurately reflects realistic ambient lighting. Finally, \sysname's tight integration with mobile AR can open new opportunities for versatile solutions in digital production, cinematic content creation, and immersive storytelling.

\para{Implications for Privacy and Security.}
Our work demonstrates the feasibility and effectiveness of generating high-quality environment lighting information from limited camera observations.
While this capability can significantly enhance the realism of virtual object rendering, it also introduces potential risks to user spatial privacy, as the reconstructed environment maps may inadvertently expose details about the user's physical surroundings.
For example, recent work demonstrates that sensitive user information can inadvertently be captured by advanced lighting estimation systems and leaked in AR streaming applications~\cite{zhao2022privacy}.
Additional types of information, such as users' location, may be inferred from the completed LDR environment map~\cite{Guzman2021-wg}.
Future work should consider defense mechanisms to prohibit the leakage of sensitive user spatial privacy information.

\para{Limitations and Future Directions}.
While \sysname demonstrates strong performance in generating high-quality environment lighting, it can be further improved in two main ways. First, a useful extension of \sysname would be to support spatially variant lighting estimation, where lighting conditions differ across physical locations. Supporting this capability requires two key changes to the lighting estimation workflow: \1 enabling 3D reconstruction of the surrounding environment to allow spatial transformation and alignment of environment observations, and \2 training generative lighting models on spatially variant environment map datasets, which often contain diverse distortion patterns. Existing solutions, such as the near-field reconstruction method in LitAR~\cite{zhao2022litar}, can be leveraged to develop 3D environment reconstruction methods. However, the primary challenge for training such generative models lies in the lack of high-quality datasets that capture spatially varying lighting. Future research could explore new data curation pipelines or integrate 3D transformation operations directly into generative models.
Second, \sysname's temporal consistency mechanism can be extended beyond the color appearance refinement to better support long AR sessions and scenes with dynamic lighting changes.
Future work may investigate methods for progressive environment map updates to enhance consistency and realism across time.

\section{Conclusion}

In this work, we introduced a lighting estimation framework \sysname that can be easily integrated into many existing mobile AR applications.
Our novel two-step generative lighting estimation pipeline ensures high-quality and robust results under multiple environmental lighting conditions, including various light intensities and color temperatures.
Our design uses pre-trained large generative models and AR context data to generate HDR environment maps more accurately from limited environment observations.
Our real-time refinement steps enhance the quality of lighting and improve responsiveness to estimation.
Comprehensive quantitative evaluation and user study confirm the accuracy and robustness of \sysname compared to recent generative models.
\begin{acks}
We thank the anonymous reviewers for their constructive feedback.
This work was supported in part by NSF Grants \#2105564, \#2236987, \#2346133, \#2350189, \#2402383, and a VMware grant.
\end{acks}

\balance
\bibliographystyle{ACM-Reference-Format}
\bibliography{main}

\newpage
\appendix

\section{Lighting Property Measurement Details}
\label{sec:additional_lighting_property_measurement_details}

We describe additional details for the environmental lighting conditions measurement study described in \S\ref{sec:preliminary}.
Our measurement focuses on two properties of environmental lighting conditions: \emph{light intensity} and \emph{color temperature}, representing an environment's overall brightness and color appearances.
We choose these two lighting condition properties because of their significant impact on virtual object rendering of all material kinds~\cite{zhao2022litar}
We select two standard lighting estimation data sources: the \emph{Laval dataset}~\cite{Gardner2017}, an academic open research dataset, and the \emph{PolyHaven} website~\cite{polyhaven}, a royalty-free HDR environment map data website.
In total, we collected 2235 and 196 data items from the Laval dataset and PolyHaven, respectively.
Following~\cite{Phongthawee2023DiffusionLight,wang2022stylelight}, we perform a standard data preprocessing procedure of a color correction process with gamma correction with $\gamma=2.4$ and setting the 99th percentile of pixel intensity to 0.9.

\begin{figure}[t]
    \centering

    \begin{minipage}{0.9\linewidth}
        \centering
        \includegraphics[width=\linewidth]{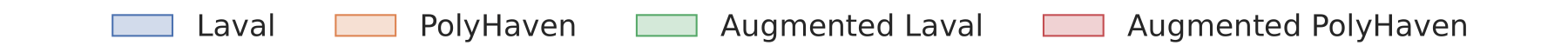}
    \end{minipage}

    \vspace{.5em}  \begin{subfigure}[b]{0.425\columnwidth}
        \centering
        \includegraphics[width=\linewidth]{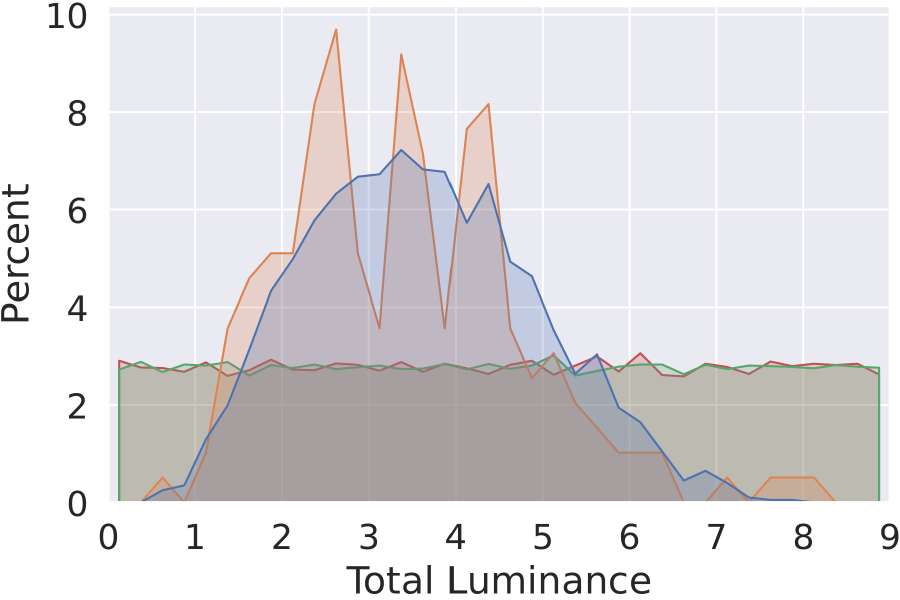}
        \caption{Light Intensity}
        \label{subfig:color_intensity}
    \end{subfigure}\quad
    \begin{subfigure}[b]{0.425\columnwidth}
        \centering
        \includegraphics[width=\linewidth]{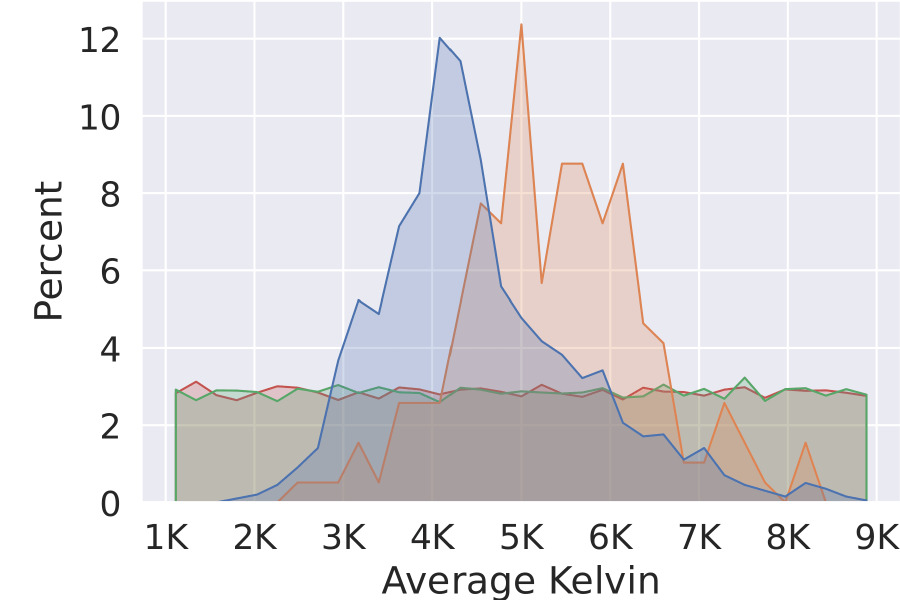}
        \caption{Color Temperature}
        \label{subfig:color_temperature}
    \end{subfigure}
\caption{
        Lighting condition measurement.
        \textnormal{We measure the distributions of light intensity (\ref{subfig:color_intensity}) and color temperature (\ref{subfig:color_temperature}) on environment maps collected from the Laval indoor dataset and PolyHaven and their augmented variants. Compared to the original data, our augmentation technique effectively creates greater diversity in the lighting features of the dataset.}
    }
\label{fig:preliminary_appendix}
\end{figure}

Next, we calculate the properties of the lighting condition.
We calculate the \emph{light intensity} as the total luminance of a given HDR environment map image.
To do so, we first calculate the individual pixel luminance $l$, which converts the original environment map image from RGB color space to the CIE XYZ color space~\cite{ibraheem2012understanding} and then derives the pixel luminance component.
We derive the total image luminance by summing the pixel luminance $l$ weighted by their differential solid angles ${\rm d}\omega_i$ throughout the HDR environment map image:

\vspace{-.5em}
\begin{equation}
    L = \sum^{N}_{i=1} (0.212671 R_i + 0.71516 G_i + 0.072169 B_i) {\rm d}\omega_i
    \label{eq:total_lumen}
\end{equation}

where $i$ represents the i-th pixel in the environment map with $N$ pixels.
For color temperature value extraction, we first calculate the average pixel RGB value from each environment map and then calculate its correlated color temperature using a recent method~\cite{ohno2014practical}.
The calculated environment map image color temperature values are given in Kelvin.
In Figure~\ref{fig:preliminary_appendix}, we visualize the measured lighting condition distributions.
While both datasets contain a wide selection of lighting conditions, the data distribution shows several biases.
Predominantly, neutral lighting conditions of light intensity and color temperatures are seen in two datasets.
For light intensity, more low-light environments than bright-lighting conditions data items are seen in the Laval dataset.
As for the color temperature distribution, a significant part of the PolyHaven data items lies in the cool color range.
These data distribution biases have also led to biased training and evaluation of lighting estimation systems.

We also evaluated the augmented versions of the Laval and PolyHaven datasets generated using our data augmentation technique (\S\ref{subsubsec:appendix_robustness_evaluation_data_generation}). The results demonstrate that our augmentation method successfully produces new dataset variants with more diverse and evenly distributed lighting characteristics. These augmented datasets are well-suited for assessing the robustness of lighting estimation systems under a wide range of lighting conditions.

\section{Dataset Generation Details}
\label{subsec:dataset_generation}

Below, we describe how we construct the training datasets for the LDR environment map completion and high-intensity map estimation tasks, from three existing open-source datasets.

\subsection{LDR Environment Map Completion Data}
We first collect a large set of LDR environment map images from two large LDR indoor environment map sources: the Matterport3D dataset~\cite{Matterport3D} and Structured3D datasets~\cite{zheng2020structured3d}.
The Matterport3D dataset provides a large set of real-world captured environment maps, and the Structured3D dataset provides a large set of synthetic environment map images with photorealistic visuals.
The combined training dataset consists of 29461 data items, \emph{10X} larger than the Laval dataset~\cite{Gardner2017}.

Next, we mask the LDR environment map images at random angles to generate AR camera observation images.
The masks are generated using the pin-hole camera model to simulate the real-world partial environment observations.
We also combine multiple image masks to simulate multi-view environment observations.
Specifically, the number of views is randomly chosen from 1 to 5, and the camera's horizontal FoV is randomly chosen between 60 and 120 degrees.
For environment semantics, we use a pre-trained semantic map estimation model~\cite{xiao2018unified} to estimate semantic maps directly from the collected LDR environment maps.
We also use the LDR image mask to generate masked semantic maps representing the environment's semantic information received by AR devices.
For the ambient lighting condition prompt generation, we create the ambient light property labels using the masked partial environment map image pixel values and our defined system threshold values in \S\ref{subsubsec:ldr_environment_map_completion}.

\subsection{High-intensity Map Estimation Data}
Our novel learning objective enables the high-intensity map estimation model to leverage its pre-trained knowledge on LDR images, allowing it to be effectively fine-tuned using a limited amount of HDR environment map data. This capability is crucial for training with the small yet high-quality HDR dataset provided by the Laval dataset.
Since the original Laval dataset contains HDR environment map images, we applied Equation~\ref{eq:hdr_scaling} to convert them into 1,489 paired samples of LDR environment maps and their corresponding high-intensity maps.
Specifically, we first clamp the raw HDR environment map to the range $[1, +\infty]$ to isolate pixel intensities beyond the LDR range, and then rescale the values to $[0, +\infty]$ to obtain the high-intensity map $I_i$.
Next, we convert the range of the raw high-intensity $[0, +\infty]$ to the standard LDR pixel range in $[0, 1]$.
Note that although this transformation is not lossless, our evaluation suggests minimal impact on the overall lighting estimation.
In addition to the generated image pair data, we add the text prompt \textbf{P2} during training.

\subsection{Robustness Evaluation Data Generation}
\label{subsubsec:appendix_robustness_evaluation_data_generation}
Evaluating the robustness of lighting estimation systems is particularly challenging because it requires controlled lighting condition changes.
To avoid costly real-time data capturing, we propose an image editing-based method that creates variants of standard lighting estimation testing datasets to represent diverse lighting conditions while maintaining the original environment map visual context.
Specifically, our method includes the following three steps.
First, we generate a set of edited variants of the Laval dataset by applying a uniform scaling term $s$ to all environment map images.
Then, we measure the total light intensity and average color temperatures using the measurement method introduced in \S\ref{sec:additional_lighting_property_measurement_details}.
Finally, we uniformly sample from the generated data to ensure equal representation of each edited environment map variant across different light intensity and color temperature ranges.
The scaling term $s$ are selected from $[0.25, 4]$ using a step of $0.125$.
For the light-intensity editing, the scaling term is uniformly applied to all three color channels, while only red and blue channels are scaled with $s$ and $1/s$ for color temperature editing~\cite{afifi2020deep}.
Figure~\ref{fig:primer_data_aug} shows examples of the edited environment maps.
We created a set of augmented Laval indoor datasets using this method, with edited average light intensities and color temperatures in the ranges of $[-20\%, 20\%]$ and $[-50\%, 50\%]$.
In total, we generated approximately 60,000 data items from the 290 original environment maps in the test split of the Laval dataset.

\section{Model Training Details}
\label{subsubsec:appendix_model_training}

Our generative lighting estimation pipeline consists of three ControlNet~\cite{zhang2023adding} models, two for LDR environment map completion with RGB and semantics context, and one for high-intensity map estimation.
We train all the models by fine-tuning the pre-trained \texttt{StableDiffusion 1.5 inpaint}~\cite{rombach2021highresolution} checkpoint.
We apply several data augmentation techniques to the previously generated datasets to train the models and improve generalization in different lighting conditions and environmental contexts.
We augment the lighting condition by applying a random scaling $s$ similar to the scaling term used in the robustness test data generation.
This augmentation is only applied to LDR environment map completion training.
For environment context augmentation, we apply horizontal rotations to environment map images.
This augmentation is used for both LDR completion and high-intensity map estimation training.
Our generative lighting estimation models are trained on affordable commercial PC hardware, specifically, a high-end workstation PC with an I9-13900K CPU and an RTX4090 GPU.
The LDR completion model requires an average of 12 hours of training, while the high-intensity map estimation model, due to its smaller training dataset, only requires an average of 4 hours.

\section{On-device Real-time Refinement Details}
\label{sec:on_device_real_time_refinement_details_appendix}

\begin{figure*}[t]
\centering
    \includegraphics[width=\linewidth]{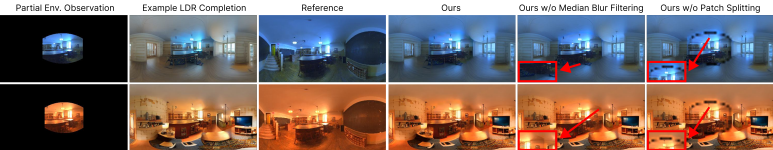}
    \caption{
        Examples of color appearance refinement.
        \textnormal{We use examples of extreme lighting conditions, extremely cool temperature (row 1) and extremely warm temperature (row 2), to show how our color refinement algorithm can improve the estimation result color accuracy. Compared to the original LDR completion result, our refined environment map color is closer to the original full environment map (marked as Reference). Columns 5 and 6 show additional results to visualize the effects of median blur filtering and patch splitting on the refinement results. Particularly, visual artifacts can be observed on the image regions between the transitioning edges of the observed and unobserved environments.}
    }
    \label{fig:color_apperance_matching}
\end{figure*}

In this section, we show additional results and technical details of the color appearance refinement technique in \sysname.
In the refinement process, our technique seeks to create a color refinement matrix of pixel color multipliers for the completed LDR environment maps.
The color refinement matrix consists of two types of multiplier values: \1 the global color multiplier and \2 the local color multiplier.
The global color multiplier adjusts the overall image colors of completed environment map images, while the local color multiplier adjusts fine-grained colors on the observed environment map regions.
The two color refinement terms are combined into a color refinement matrix and applied to the completed LDR environment maps.

\begin{figure*}[t]
\centering
    \includegraphics[width=\linewidth]{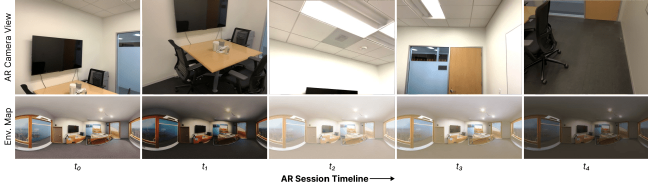}
    \caption{
        Examples of environment map refinement over an AR session.
        \textnormal{We demonstrate the effectiveness of our refinement step using selected frames from an AR session in the ARKitScene~\cite{dehghan2021arkitscenes} dataset. In this example, the initial environment map is estimated by \sysname at time $t_0$, and subsequently refined at $t_1$ through $t_4$ using the corresponding AR camera images. As the AR session progresses, the ambient lighting intensity varies. The color refinement step in \sysname effectively adapts the estimated environment maps to reflect these changing lighting conditions. Note that the original AR camera video has been lightly edited to enhance intensity variations for improved visual clarity.}
    }
    \label{fig:refinement_over_time}
\end{figure*}

Our technique calculates the global color multiplier by deriving per-channel multipliers as the ratio between average colors in the estimated and observed environment map images.
The local multipliers are calculated by first splitting the estimated environment map images and the partial environment observation image into $NxM$ patches.
Then, our technique calculates local multipliers as the average color ratios for each patch.
A 3x3 median blur filter is used to address color smoothness between the observation edges.
Empirically, we found 8x8 is the best image patch size.

We show the effectiveness of our refinement technique via two visual quality evaluation.
As shown in Figure~\ref{fig:color_apperance_matching}, our refinement technique can adjust environment map colors even in very challenging lighting conditions.
Furthermore, our design choices of median blurring and patch splitting are useful to smooth out the artifacts on the edges.
Figure~\ref{fig:refinement_over_time} illustrates the effectiveness of this refinement process over time during an AR session. The example video is sourced from the ARKitScene~\cite{dehghan2021arkitscenes} dataset, with minor edits applied to enhance lighting intensity variations. \sysname first estimates the environment map at frame $t_0$, and subsequently refines it at frames $t_1$ through $t_4$ using the corresponding AR camera images. The refined environment maps clearly reflect the observed changes in ambient lighting intensity. This refinement step is essential for maintaining coherence between the physical environment and the rendered virtual content as lighting conditions change over time. In practice, it is especially effective for adapting to minor lighting changes that do not involve significant alterations to scene geometry or object placement.

\end{document}